\documentclass[journal]{IEEEtran}
\IEEEoverridecommandlockouts
\usepackage{amsmath} 
\usepackage{amssymb}  
\usepackage{bm}
\usepackage{siunitx}
\usepackage{xcolor}
\usepackage{multirow} 
\usepackage{graphicx} 
\usepackage{url}
\usepackage[hidelinks]{hyperref}
\usepackage{booktabs}
\usepackage{subcaption}
\usepackage{algorithm}
\usepackage{algpseudocode}

\newenvironment{keywords}{\begin{IEEEkeywords}}{\end{IEEEkeywords}}

\title{\textbf{PDM}: Improved Adversarial Imitation Learning for Legged Robots via \textbf{P}arallel-Policy \textbf{D}istribution \textbf{M}atching }
\author{Jiahui Zhang$^{*1}$, Zeren Luo$^{*1}$, Yimin Han$^{*1}$,\\ Yeke Chen$^{1}$, Shihao Dong$^{1}$, Liu Zhao$^{1}$, Yudong Zhao$^{2}$, Annan Tang$^{2}$ and Peng Lu$^{\dagger1}$
\thanks{$^{*}$ Equal Contribution. $^{\dagger}$Corresponding author}
\thanks{$^{1}$ Adaptive Robotic Controls Lab (ArcLab), Department of Mechanical Engineering, University of Hong Kong, Hong Kong SAR, China, \url{holmesz, zerluo@connect.hku.hk,lupeng@hku.hk}.}
\thanks{$^{2}$ EngineAI Robotics Inc., Guangdong, China, \url{zhaoyd, robert_nan@engineai.com.cn}.}}

\IEEEaftertitletext{%
  \begin{minipage}{\textwidth}
    \centering
    \includegraphics[width=0.99\textwidth]{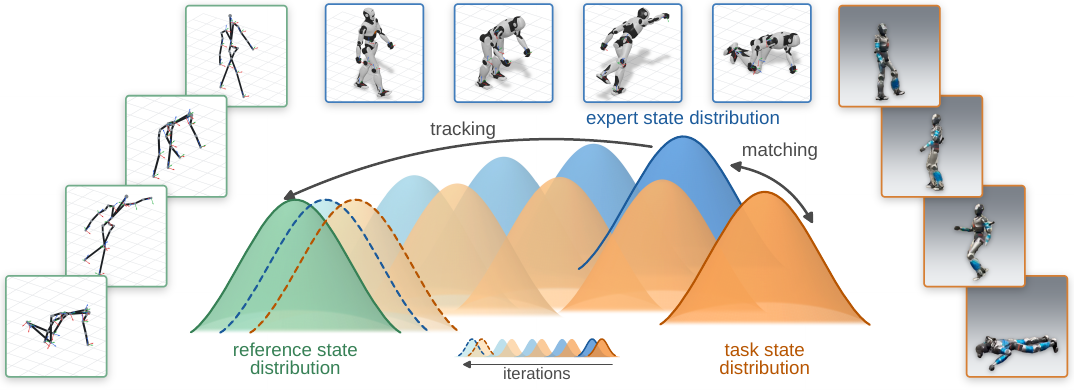}
    \captionof{figure}{Concept of PDM: the tracking reward moves the expert state distribution toward the reference state distribution, while distribution matching draws the state distribution of the downstream task policy toward the evolving expert state distribution over the training iterations.}
    \label{fig:cover}
  \end{minipage}%
  \vspace{1em}%
}
\begin{document}

\maketitle

\begin{abstract}
Adversarial imitation learning enables agents to acquire skills from expert demonstrations, but early discriminator saturation can leave policies with uninformative rewards.
We introduce Parallel-Policy Distribution Matching (PDM), a multi-task
reinforcement learning framework with separate tracking and downstream
policies. Both policies are trained concurrently from scratch with
distinct tracking and task objectives.
The tracking policy follows reference motions and generates physically
feasible trajectories online. Stochastic rollouts from both policies
provide samples for discriminator training.
We couple the policies through occupancy-measure matching and derive
discriminator-based matching rewards for both.
Simulation experiments show that the generated trajectories improve
downstream tracking accuracy and that stochastic training samples help
mitigate early discriminator saturation.
In real-world experiments, the downstream policy preserved the desired
motion style during locomotion, recovered from disturbances, climbed
obstacles of varying heights, and executed a flip on command.
For a brief introduction and videos, see~\url{https://p2dm.github.io/}
\end{abstract}

\section{INTRODUCTION}

Humanoid robots are being developed for factories, warehouses, homes,
and public environments designed around people. Their human-like form
is useful only if they can perform the corresponding whole-body skills.
Reinforcement learning with hand-crafted objectives can produce agile
behavior in massively parallel simulators~\cite{rudin2022learning, luo2025mild}.
Each new skill requires an appropriate objective, however, and motion
qualities such as naturalness are difficult to specify explicitly.
Human motion data provide an alternative source of supervision.
Two common paradigms use these data differently: feature-based tracking matches a time-aligned reference frame by frame, whereas adversarial imitation matches a state distribution.

Feature-based tracking was established by
DeepMimic~\cite{peng2018deepmimic}. This approach derives dense rewards
from framewise errors in kinematic features and now supports dynamic
skills on physical
humanoids~\cite{liao2025beyondmimic,cheng2024exbody,he2024omnih2o,chen2025gmt,ze2025twist}.
The dense signal encourages accurate reproduction. Conventional
tracking policies are conditioned on a reference trajectory or phase,
although task changes may require them to depart from that trajectory.
Existing methods relax this dependence through phase fallback under
conflicting task constraints~\cite{sleiman2024mcp} or sparse keyframe
tracking~\cite{zargarbashi2024robotkeyframing}. Others use constraint-gated
imitation~\cite{wen2025consmimic} or omit reference inputs at
deployment~\cite{sood2025apex}.
These methods primarily use tracking supervision to shape the controller
itself. Our framework uses tracking only during training to generate
motion supervision for a separate downstream policy.

Adversarial imitation removes framewise alignment and treats imitation
as distribution matching. Generative adversarial imitation learning
(GAIL)~\cite{ho2016generative} matches expert and policy occupancy
measures, using a discriminator to provide the reward.
The adversarial motion priors (AMP) framework~\cite{peng2021amp} applies this
supervision to short motion transitions. Its discriminator output is a
style reward that can be combined with task
rewards~\cite{escontrela2022adversarial, luo2025learning}.
Meta Motivo~\cite{tirinzoni2025zero} interprets this reward at the
distribution level. It writes the Kullback--Leibler (KL) divergence
between a policy's state distribution and that of an unlabeled motion
dataset as an expected return. The per-step reward is the log density
ratio of the two distributions, estimated by the discriminator logit.
BFM-Zero~\cite{li2026bfm} uses the same reward to regularize whole-body
control policies for a real humanoid.
Under this interpretation, the reward measures how close the policy
distribution is to the dataset distribution, while the logit indicates
how well they match. When the distributions coincide, the logits of
dataset and policy samples converge to a common equilibrium value.
Persistent separation between these logits indicates discriminator
saturation.

Prior work has addressed instability, discriminator saturation, and
mode collapse through Wasserstein
objectives~\cite{li2023learning,tang2024humanmimic} and latent-conditioned
representations~\cite{peng2022ase,tessler2023calm,dou2023case,vollenweider2023advanced}.
These methods modify the adversarial objective or motion representation
but retain the dataset as the target distribution. They do not ensure
that reference transitions used as positive samples are consistent with
the robot dynamics.
Kinematic retargeting is a common source of this mismatch. It maps
human motion to the robot before policy training and can introduce foot
sliding, ground penetration, and violations of actuation
limits~\cite{araujo2025retargeting,yang2025omniretarget}.
Some retargeted transitions may therefore be inconsistent with the
dynamics and contacts used to optimize the policy. The discriminator
can exploit these artifacts to separate reference and policy samples
instead of relying on behaviorally meaningful differences.
The logits of reference and policy samples can then remain separated,
causing early discriminator saturation and an uninformative style reward.
The target distribution supplying positive samples therefore matters
alongside the adversarial objective.

Existing remedies fall into three families with different costs.
The first changes how imitation influences policy optimization through
Wasserstein rewards~\cite{li2023learning,tang2024humanmimic}, task-priority
gradient projection~\cite{zhang2026tcmp}, or constraint-dependent
gating~\cite{wen2025consmimic,xu2026selectiveamp}. The underlying reference
transitions remain unchanged.
The second refines retargeted data through kinodynamic or physics-aware
optimization~\cite{wang2026spark,ferrari2026dynaretarget,muller2026reactor},
adding a motion-processing stage tied to a chosen dynamics model.
The third uses rollouts from a trained teacher as executable supervision.
For example, adversarial distillation~\cite{long2025styleloco} first
trains a teacher, then uses its behavior to supervise a student.
These staged methods provide robot-generated data, but the teacher is
trained before the downstream policy and is not updated jointly with it.

We introduce \textbf{Parallel-Policy Distribution Matching (PDM)}, a
framework that combines feature-based motion tracking with motion
imitation based on a generative adversarial network (GAN).
PDM concurrently trains a tracking expert policy and a downstream task
policy with KL-divergence regularization. Each policy pursues its own
tracking or task objective.
The expert receives dense motion-tracking rewards and generates
physically feasible trajectories in simulation.
We train the discriminator on stochastic rollouts from both policies.
Its distribution-matching reward accounts for all input features when
distinguishing samples from the two policies.
We observed that the downstream policy generated samples that challenged
the discriminator. These samples contributed to a smaller logit gap and
better preservation of the desired motion style during downstream task
execution on hardware.
The contributions are as follows:
\begin{itemize}
    \item We offer a new perspective on understanding and shaping
    discriminator behavior through the distribution of samples in its
    training buffer. Stochastic rollouts prevent
    discriminator overfitting, while excessive action noise makes the
    matching reward less informative.
    \item We extend occupancy-measure matching to concurrent training
    of a tracking policy and a downstream policy. We show that a single
    discriminator provides matching rewards for both.
    \item We evaluate PDM in simulation and real-world experiments.
    The discriminator logits converge without early saturation in
    simulation, and the downstream policies robustly execute diverse
    tasks with the desired motion style on hardware.
\end{itemize}

\section{RELATED WORK}

\subsection{Feature-Based Motion Imitation and Tracking}
DeepMimic~\cite{peng2018deepmimic} popularized dense, phase-aligned
feature matching for physics-based motion imitation.
This approach now supports whole-body humanoid tracking on
hardware~\cite{cheng2024exbody,he2024omnih2o,fu2024humanplus,liao2025beyondmimic,chen2025gmt,ze2025twist}
and structured motion representations that scale to large clip
collections~\cite{luo2023phc,luo2023pulse,tessler2024maskedmimic,serifi2024vmp}.
Related work relaxes exact tracking when task constraints require
deviations or improves motion data through physics-based
augmentation~\cite{sleiman2024mcp,zargarbashi2024robotkeyframing,wen2025consmimic,xu2025parc}.

These methods use tracking supervision to train the controller itself
or learn a representation that is later distilled.
In PDM, a separate tracking policy executes transitions in simulation
to provide adversarial supervision. This policy is used only during
training. The deployed downstream policy receives neither a reference
frame nor a phase input.

\subsection{Adversarial Motion Priors and Latent Structure}
GAIL~\cite{ho2016generative} formulates imitation as adversarial matching
of expert and policy occupancy measures over state--action pairs.
AMP~\cite{peng2021amp} applies adversarial supervision to short
motion-transition features. AMP-style discriminators have since been
used as motion priors for robotic
control~\cite{escontrela2022adversarial,vollenweider2023advanced}.
Later work improves coverage and controllability through latent or
behavior-conditioned
representations~\cite{peng2022ase,tessler2023calm,dou2023case,tirinzoni2025zero}.
Wasserstein objectives can reduce early
saturation~\cite{li2023learning,tang2024humanmimic}.
A recent survey examines trade-offs between feature-based and GAN-based rewards and
discusses structured motion representations~\cite{li2025featuregan}.

These formulations generally train the discriminator against a motion
dataset fixed before policy optimization. Latent structure addresses
coverage and controllability within that dataset. PDM changes the source
and update schedule of the discriminator's reference samples.
The two directions are complementary: PDM does not introduce a latent
skill representation, and latent conditioning does not ensure the
dynamic consistency of retargeted transitions.

\subsection{Imperfect and Dynamically Inconsistent References}
Work on imperfect demonstrations modifies either the learning objective
or the reference data. For the objective, Wasserstein rewards retain a
usable signal from a strong
discriminator~\cite{li2023learning,tang2024humanmimic}. Task-priority
gradient projection protects the task objective from conflicting imitation
gradients~\cite{zhang2026tcmp}, while gated imitation applies tracking
or style rewards only when they are locally compatible with the
task~\cite{wen2025consmimic,xu2026selectiveamp}.
For reference data, retargeting studies show that quality materially
affects downstream
tracking~\cite{araujo2025retargeting,yang2025omniretarget}.
Kinodynamic trajectory optimization refines kinematic clips under a
dynamics model~\cite{wang2026spark,ferrari2026dynaretarget}, and bilevel
methods jointly adapt retargeting parameters and a tracking
policy~\cite{muller2026reactor}.

Objective-level methods can reduce the effect of a poor reference but
do not replace its transitions. Data-refinement methods address
reference quality through an additional optimization stage tied to the
selected dynamics model.
PDM trains the discriminator on transitions executed by the tracking
policy under sampled dynamics in the randomized training simulator.
The joint data-adaptation method in~\cite{muller2026reactor} optimizes a
retargeting map together with a tracking policy. PDM updates the realized
reference distribution online and uses it to supervise a separate
downstream policy.

\subsection{Self-Imitation and Staged Supervision}
Self-imitation learning~\cite{oh2018self} and generative
adversarial self-imitation learning~\cite{oh2018gasil} reuse
high-return trajectories generated by the learning agent.
Hindsight generative adversarial imitation
learning~\cite{guo2019hgail} uses hindsight transformations to synthesize
expert-like samples.
The unlabeled imperfect demonstrations method~\cite{wang2023uid}
selects samples from a fixed imperfect-demonstration set according to
their compatibility with the evolving policy.
Self-supervised adversarial imitation
learning~\cite{monteiro2023sail} combines teacher observations with
selected policy rollouts for self-supervised imitation.
In these methods, supervision comes from the learning policy's own
behavior or is selected for compatibility with it.

StyleLoco~\cite{long2025styleloco} uses policy-generated references in
stages. It first trains a task-oriented control policy, then trains the
deployable policy with separate discriminators for the first policy's
rollouts and for motion-capture data. The first policy is trained
beforehand and is not updated jointly with the deployable policy.

PDM also uses policy-executed transitions as adversarial positive
samples. They come from a separate tracking policy optimized
concurrently with the downstream policy. The tracking policy remains
anchored to the motion reference, while a single discriminator compares
the two rollout distributions as the reference distribution evolves.
\begin{figure*}[t]
    \centering
    \includegraphics{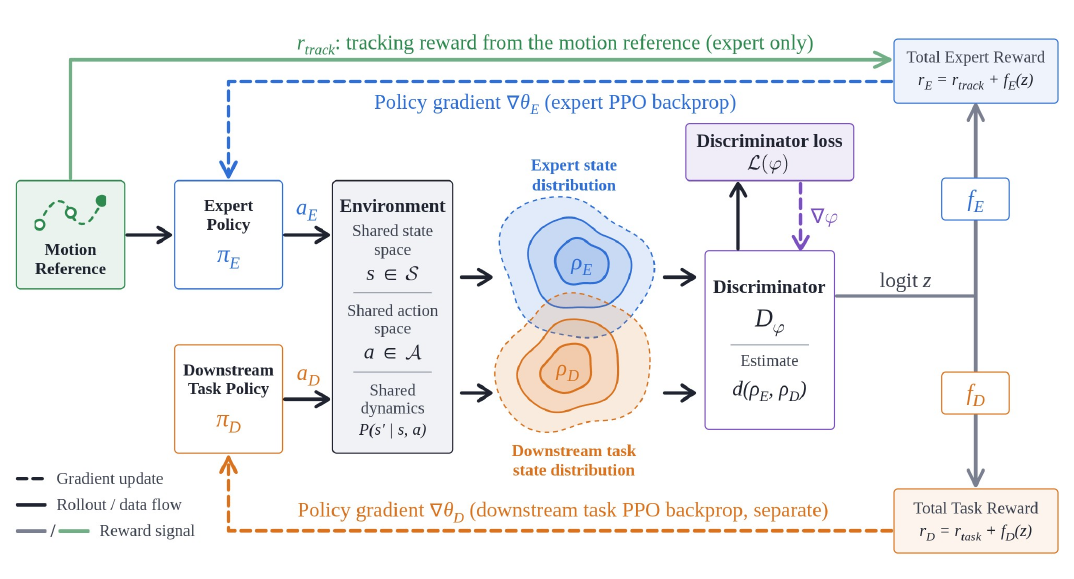}
    \caption{Overview of PDM. The
    expert policy $\pi_E$ and the downstream policy
    $\pi_D$ interact with the same randomized robot dynamics and
    action space but retain separate MDPs,
    parameters, and PPO updates. A
    single discriminator $D_{\varphi}$ is trained to separate the expert
    distribution $\rho_E$ from the downstream distribution $\rho_D$, and
    its logit $z$ is mapped by $f_E$ and $f_D$ to the
    distribution-matching rewards of the two branches.}
    \label{fig:PDM_overview}
\end{figure*}
\subsection{Multi-Task Concurrent Training}
A smaller group of methods partitions parallel environments between
complementary training modes.
Concurrent teacher-student reinforcement
learning~\cite{jia2024cts} trains privileged teacher and proprioceptive
student groups concurrently with shared policy and critic networks.
The partition separates information pathways for one locomotion
objective.
Hybrid imitation learning (HIL)~\cite{wang2025hil} trains one controller
in frame-level tracking and AMP-based target-following modes, with a
unified observation space after a tracking warm-up.
HIL is structurally close to PDM, but its discriminator draws positive
samples from a fixed, physics-refined reference-motion dataset. PDM
uses online rollouts from the tracking environments.
Generalizing from References~\cite{wang2026gfr} jointly trains a
single goal-conditioned policy on reference-guided imitation and
goal-reaching tasks without an adversarial objective.

PDM assigns separate policies and critics to the two environment groups.
The tracking group optimizes reference tracking, while the downstream
group pursues its task objective. Transitions executed by the tracking
group define the discriminator reference.
This separation is essential to PDM: the tracking objective shapes the
source distribution without becoming an observation or a deployment-time
constraint for the downstream policy.

\section{Method}
\label{sec:method}
Section~\ref{sec:parallel_mdp} defines the two learning problems, the
notation, and the overall objective.
Section~\ref{sec:distribution_objective} decomposes this objective into
two policy objectives, each of which maximizes return with a
Kullback--Leibler (KL) regularizer on the state distributions.
Section~\ref{sec:adversarial_matching} uses an adversarial discriminator
to estimate this regularizer as a reward, and
Section~\ref{sec:parallel_training} describes the training procedure.
Figure~\ref{fig:PDM_overview} gives an overview of the framework.

\subsection{Problem Formulation with Parallel MDPs}
\label{sec:parallel_mdp}
A multi-task reinforcement-learning problem is commonly formulated as a
family of Markov decision processes (MDPs) with a shared policy or shared
parameters. We train two separate policies, each in its own MDP, with
shared dynamics and a shared action space. Let $s\in\mathcal{S}$ denote
the robot state and
$a\in\mathcal{A}$ the action. Both policies use the transition dynamics
$s'\sim P(\cdot\mid s,a)$. The two MDPs are
$\text{MDP}_E=(\mathcal{S},\mathcal{A},P,r_E,\gamma_E,\mu_E)$ for the expert
policy and $\text{MDP}_D=(\mathcal{S},\mathcal{A},P,r_D,\gamma_D,\mu_D)$ for
the downstream policy, where $r_i$ is the per-step reward,
$\gamma_i\in[0,1)$ is the discount factor, and $\mu_i$ is the initial-state
distribution for $i\in\{E,D\}$.
Although the two MDPs share $\mathcal{S}$, $\mathcal{A}$, and $P$, their
policies receive different goals and observations. The expert policy
$a\sim\pi_E(\cdot\mid\bm o_E)$ is conditioned on a tracking goal, usually
the states of the reference motion, whereas the observation
$\bm o_D$ of the downstream policy $a\sim\pi_D(\cdot\mid\bm o_D)$ usually
contains a task-dependent command.

Distribution matching couples the policies through their induced state
distributions. Let $\rho_E$ and $\rho_D$ denote the
distributions of $s_t$ induced by rolling out $\pi_E$ in $\text{MDP}_E$ and
$\pi_D$ in $\text{MDP}_D$, respectively. We write $s_E\sim\rho_E$ and
$s_D\sim\rho_D$ for the corresponding samples, and $s_{E,t}$ and $s_{D,t}$
for the states at time step $t$ of expert and downstream rollouts,
respectively. The overall objective maximizes the expected return of both
policies while penalizing the divergence between $\rho_E$ and $\rho_D$:
\begin{equation}
\begin{aligned}
\max_{\pi_E,\pi_D}\quad
&\mathbb{E}_{i\sim p(i)}
\left[
\mathbb{E}_{\tau_i\sim(\mathrm{MDP}_i,\pi_i)}
\left[
\sum_{t=0}^{\infty}\gamma_i^t r_{i,t}(s_t,a_t)
\right]
\right]
\\
&\quad - \beta\,
\mathrm{KL}_\mathrm{sym}\!\left(\rho_E \,\middle\|\, \rho_D\right),
i\in\{E,D\},
\end{aligned}
\label{eq:max_return}
\end{equation}
where $\tau_i$ is a trajectory generated by $\pi_i$ in $\text{MDP}_i$,
$p(i)$ is the probability of assigning an agent to $\text{MDP}_i$, and
$\mathrm{KL}_\mathrm{sym}(\rho_E\Vert\rho_D)
=\mathrm{KL}(\rho_E\Vert\rho_D)+\mathrm{KL}(\rho_D\Vert\rho_E)$. 

The tracking expert in PDM generates motion samples online for
discriminator training, replacing the fixed reference dataset used in
conventional adversarial imitation learning. This design enables
distribution-matching regularization for both policies.
Because $\rho_E$ is induced by executing $\pi_E$ under the dynamics $P$,
it contains only states realized by the simulated robot. This differs
from the kinematically retargeted motion distribution derived from
$\mathcal{M}$.

\subsection{Decomposition into Parallel Subproblems}
\label{sec:distribution_objective}
We decompose the objective in \eqref{eq:max_return} into two
reinforcement-learning subproblems and solve them in parallel. Each
subproblem maximizes the return of one policy, with the objectives
\begin{subequations}
\label{eq:distribution_regularized_objectives}
\begin{align}
    J_E(\pi_E)
    ={}&
    \mathbb{E}_{\substack{
        \tau_E\sim\pi_E,\\
        s_0\sim\mu_E
    }}
    \left[
        \sum_{t=0}^{\infty}
        \gamma_E^t r_t^{\mathrm{track}}
    \right]
    \notag\\
    &-
    \beta_E
    \mathrm{KL}
    \left(
        \rho_E
        \,\Vert\,
        \rho_D
    \right),
    \label{eq:executable_return}
    \\
    J_D(\pi_D)
    ={}&
    \mathbb{E}_{\substack{
        \tau_D\sim\pi_D,\\
        s_0\sim\mu_D
    }}
    \left[
        \sum_{t=0}^{\infty}
        \gamma_D^t r_t^{\mathrm{task}}
    \right]
    \notag\\
    &-
    \beta_D
    \mathrm{KL}
    \left(
        \rho_D
        \,\Vert\,
        \rho_E
    \right).
    \label{eq:distribution_regularized_objective}
\end{align}
\end{subequations}
The two policies pursue distinct rewards, with regularization encouraging
their state distributions to remain close.
The expert policy tracks the reference motion \(\mathcal{M}\) using the
tracking reward \(r_t^{\mathrm{track}}\), whereas the downstream policy
pursues its task reward \(r_t^{\mathrm{task}}\) while remaining consistent
with the robot-executable and physically feasible state distribution
generated by the expert. The coefficients \(\beta_E\) and \(\beta_D\)
balance each policy's return against its KL regularizer and allow the
regularization strengths to be set separately.
Appropriate coefficients can improve policy performance. Excessive
regularization, however, can cause the policies to collapse to a
degenerate solution in which their state distributions match but neither
policy completes its task.

Each policy is regularized by the reverse KL divergence to the other
policy's distribution. This regularizer penalizes samples that are
unlikely under the other policy's distribution.
When optimizing one policy, we treat the other policy's distribution as
fixed, so neither objective introduces a gradient path between the
policies.
The return term determines \emph{what} each policy should accomplish,
whereas the KL term regularizes \emph{how} it executes the behavior.
For the downstream policy, this separation permits departures from the nominal motion when the task requires them.

\subsection{Adversarial Distribution-Matching Reward}
\label{sec:adversarial_matching}

The KL terms in \eqref{eq:executable_return} and
\eqref{eq:distribution_regularized_objective} are intractable because the
state densities are unknown and depend on the policies
through the non-differentiable dynamics $P$. 
Although the KL regularizer can be converted into a reinforcement-learning
reward and estimated with an objective similar to that of a generative
adversarial network (GAN)~\cite{tirinzoni2025zero,li2026bfm}, these methods
apply only to
one-sided distribution matching toward the fixed state
distribution of a finite reference motion $\mathcal{M}$.
We extend this conversion to bilateral distribution matching. In our
parallel-policy formulation, $\rho_E$ and $\rho_D$ are induced by
$\pi_E$ and $\pi_D$, respectively, and evolve as the policies are updated.
The discriminator therefore learns to distinguish samples from the
current policy distributions instead of comparing policy samples with a
fixed dataset. A single discriminator suffices to provide matching
rewards for both policies because it distinguishes between their samples.
We derive the result for the downstream policy; the expert case is analogous.
\subsubsection{Distribution Matching as Return Maximization}
We define $\rho_D$ and $\rho_E$ as the normalized discounted visitation
distributions of $s_t$ under $\pi_D$ and $\pi_E$, respectively. Under
these definitions, the KL term in
\eqref{eq:distribution_regularized_objective} can be written as a
negative expected return (Appendix~\ref{app:kl_derivation}):
\begin{equation}
\begin{aligned}
    &\mathrm{KL}
    \left(
        \rho_D
        \,\Vert\,
        \rho_E
    \right)
    \\
    &\quad=
    -(1-\gamma_D)
    \mathbb{E}_{\substack{
        \tau_D\sim\pi_D,\\
        s_0\sim\mu_D
    }}    
    \left[
        \sum_{t=0}^{\infty}
        \gamma_D^t
        \log
        \frac{\rho_E(s_t)}{\rho_D(s_t)}
    \right].
\end{aligned}
    \label{eq:kl_history}
\end{equation}
The log ratio inside the expectation defines the distribution-matching
reward of the downstream policy,
\begin{equation}
    \bar{r}_{D}(s)
    =
    \log
    \frac{\rho_E(s)}{\rho_D(s)},
    \label{eq:ideal_style_reward}
\end{equation}
which is positive for states that are more likely under the expert
distribution than under the downstream distribution. Absorbing the
positive constant $1-\gamma_D$ into $\beta_D$ and substituting
\eqref{eq:kl_history} into
\eqref{eq:distribution_regularized_objective} yields
\begin{equation}
    J_D(\pi_D)
    =
    \mathbb{E}_{\substack{
        \tau_D\sim\pi_D,\\
        s_0\sim\mu_D
    }}
    \left[
        \sum_{t=0}^{\infty}
        \gamma_D^t
        \left(
            r_t^{\mathrm{task}}
            +
            \beta_D
            \bar{r}_D(s_{D,t})
        \right)
    \right].
    \label{eq:augmented_objective}
\end{equation}
The KL regularizer of the downstream policy therefore becomes a per-step
reward that can be estimated after each environment step.

\subsubsection{Log-Ratio Estimation from Policy Rollouts}
Whereas prior adversarial imitation
algorithms~\cite{ho2016generative,peng2021amp,zhang2025ADD} train the
discriminator to distinguish reference motion samples from policy
samples, PDM trains it to distinguish states sampled from the two
policies. This change is the main source of the difference in
discriminator training dynamics.
We parameterize the discriminator using its logit $z_{\varphi}(s)$ and the
sigmoid function $\sigma$ as
$D_{\varphi}(s)=\sigma\left(z_{\varphi}(s)\right)$, which predicts the
probability that a state $s$ was drawn from $\rho_E$ rather than from
$\rho_D$. The discriminator is trained with the binary
cross-entropy objective
\begin{align}
    \mathcal{L}(\varphi)
    ={}&
    -
    \mathbb{E}_{s\sim \rho_E}
    \left[
        \log D_{\varphi}(s)
    \right]
    \nonumber\\
    &-
    \mathbb{E}_{s\sim \rho_D}
    \left[
        \log\left(1-D_{\varphi}(s)\right)
    \right]
    +
    \lambda_{\mathrm{GP}}
    \mathcal{L}_{\mathrm{GP}},
    \label{eq:discriminator_loss}
\end{align}
where $\lambda_{\mathrm{GP}}$ weights the gradient penalty
\begin{equation}
\begin{aligned}
    \mathcal{L}_{\mathrm{GP}}
    ={}
    \mathbb{E}_{s\sim\rho_E}
    \left[
        \left\|
            \nabla_{s}z_{\varphi}(s)
        \right\|_2^2
    \right]
    +
    \mathbb{E}_{s\sim\rho_D}
    \left[
        \left\|
            \nabla_{s}z_{\varphi}(s)
        \right\|_2^2
    \right].
\end{aligned}
    \label{eq:two_sided_gradient_penalty}
\end{equation}
We penalize the gradient of the logit
$z_{\varphi}$ on samples from both $\rho_E$ and $\rho_D$ to keep the
reward smooth in regions visited by the policies.

For fixed $\rho_E$ and $\rho_D$, omitting the gradient penalty gives the
optimal discriminator that minimizes the objective in
\eqref{eq:discriminator_loss}~\cite{goodfellow2014generative,nowozin2016f},
\begin{equation}
    D^*
    =
    \frac{\rho_E}
    {\rho_E+\rho_D}.
    \label{eq:optimal_discriminator}
\end{equation}
Solving \eqref{eq:optimal_discriminator} for the density ratio and
replacing $D^*$ with the trained discriminator yields
\begin{equation}
\begin{aligned}
    \bar{r}_D
    &=
    \log
    \frac{D^*}{1-D^*}
    \\
    &\approx
    \log
    \frac{D_{\varphi}}{1-D_{\varphi}}
    =
    \log
    \frac{\sigma(z_{\varphi})}{1-\sigma(z_{\varphi})}
    =
    z_{\varphi}.
\end{aligned}
    \label{eq:density_ratio_estimate}
\end{equation}
The discriminator logit is
therefore an estimate of the distribution-matching reward in
\eqref{eq:ideal_style_reward}, and the accuracy of this estimate
depends on how closely $D_{\varphi}$ approaches $D^*$.
For $\mathrm{KL}(\rho_E\Vert\rho_D)$ in \eqref{eq:executable_return},
the same derivation holds with the roles of the two distributions
exchanged (Appendix~\ref{app:kl_derivation}). Accordingly, the
distribution-matching reward of the expert becomes
$\bar{r}_E=\log(\rho_D/\rho_E)\approx-z_{\varphi}$,
which is the logit of the discriminator that predicts the probability
that $s$ was drawn from $\rho_D$ rather than from $\rho_E$.
Because $D_E=1-D_{\varphi}$, no separate discriminator is needed for the
expert policy: both distribution-matching rewards are obtained from the same logit $z_{\varphi}$.

\begin{figure}[htbp]
    \centering
    \includegraphics[width=0.45\textwidth]{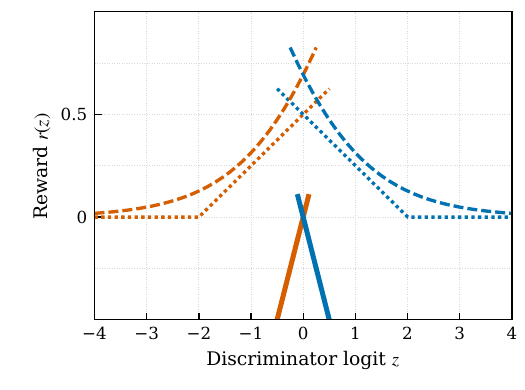}
    \caption{The discriminator logit $z$ can decrease sharply.
    Two alternative mappings therefore replace the identity mapping in
    practice and keep the reward nonnegative.
    Solid lines show $f_D$ for the
    downstream policy, and dashed lines show the corresponding
    $f_E$ for the expert policy.}
    \label{fig:discrinmiantor_reward}
\end{figure}
\subsubsection{Reward Mapping}
The raw logit $z_{\varphi}$ can decrease sharply and become negative
during training. Because negative rewards can shorten episodes, we map
the logit through $f_D$ and define the downstream
distribution-matching reward as
\begin{equation}
    r_{D,t}^{S}
    =
    f_D\left(z_{\varphi}(s_{D,t})\right).
    \label{eq:style_reward}
\end{equation}
Figure~\ref{fig:discrinmiantor_reward} compares three mappings. 
The identity
$f_D(z)=z$ recovers \eqref{eq:density_ratio_estimate}.
We use the clipped linear mapping
$f_D(z)=0.25\left(\max(z,-2)+2\right)$ to obtain a nonnegative reward.
For comparisons with AMP, we also use the logarithmic mapping
$f_D(z)=-\log\left(1-\sigma(z)\right)$, which corresponds to minimizing
the Jensen--Shannon divergence between $\rho_E$ and
$\rho_D$~\cite{ho2016generative}.
In our simulation experiments, the two
mappings produced only small differences in training logits and
tracking performance.

Replacing $\bar{r}_D$ and $\bar{r}_E$ in \eqref{eq:augmented_objective}
and its expert counterpart by the mapped rewards, with $f_E(z)=f_D(-z)$
for the expert, gives the final rewards of $\text{MDP}_E$ and
$\text{MDP}_D$:
\begin{subequations}
\label{eq:combined_rewards}
\begin{align}
    r_{E,t}
    &=
    r_t^{\mathrm{track}}
    +
    \beta_E
    f_E\left(z_{\varphi}(s_{E,t})\right),
    \label{eq:expert_combined_reward}
    \\
    r_{D,t}
    &=
    r_t^{\mathrm{task}}
    +
    \beta_D
    f_D\left(z_{\varphi}(s_{D,t})\right).
    \label{eq:downstream_combined_reward}
\end{align}
\end{subequations}
Setting $\beta_E=0$ in \eqref{eq:expert_combined_reward} reduces the
expert reward to $r_t^{\mathrm{track}}$, so the downstream policy no
longer affects the expert. The downstream reward in
\eqref{eq:downstream_combined_reward}, however, still draws the
downstream policy toward the expert's state distribution through $f_D$;
the coupling between the two policies therefore becomes one-way.

\begin{algorithm}[tbp]
    \caption{Parallel-Policy Distribution Matching}
    \label{alg:PDM}
    \small
    \begin{algorithmic}[1]
        \State \textbf{input} $\mathcal{M}$: reference motion clips
        \State \textbf{input} $\beta_E,\beta_D$: KL weights for each branch
        \State $D_{\varphi}\leftarrow$ initialize discriminator
        \State $\pi_E,\pi_D\leftarrow$ initialize policies
        \State $V_E,V_D\leftarrow$ initialize value functions
        \State $\mathcal{B}_E,\mathcal{B}_D\leftarrow\emptyset$
        initialize discriminator buffers
        \Statex
        \While{not done}
            \For{trajectory $i=1,\ldots,m$}
                \State $\tau_E^i\leftarrow$ collect trajectory in
                $\text{MDP}_E$ with $\pi_E$ tracking $\mathcal{M}$
                \State $\tau_D^i\leftarrow$ collect trajectory in
                $\text{MDP}_D$ with $\pi_D$
                \For{time step $t=0,\ldots,T-1$}
                    \State $r_{E,t}\leftarrow$ calculate reward with
                    \eqref{eq:expert_combined_reward}
                    \State $r_{D,t}\leftarrow$ calculate reward with
                    \eqref{eq:downstream_combined_reward}
                    \State record $r_{E,t}$ in $\tau_E^i$ and
                    $r_{D,t}$ in $\tau_D^i$
                \EndFor
                \State store states $s_{E,t}$ of $\tau_E^i$ in
                $\mathcal{B}_E$
                \State store states $s_{D,t}$ of $\tau_D^i$ in
                $\mathcal{B}_D$
            \EndFor
            \Statex
            \For{update step $=1,\ldots,N_{\mathrm{disc}}$}
                \State $b_E\leftarrow$ sample mini-batch of states from
                $\mathcal{B}_E$
                \State $b_D\leftarrow$ sample mini-batch of states from
                $\mathcal{B}_D$
                \State update $D_{\varphi}$ according to
                \eqref{eq:discriminator_loss} using $b_E$ and $b_D$
            \EndFor
            \Statex
            \State update $V_E$ and $\pi_E$ with PPO using
            $\{\tau_E^i\}_{i=1}^{m}$
            \State update $V_D$ and $\pi_D$ with PPO using
            $\{\tau_D^i\}_{i=1}^{m}$
        \EndWhile
    \end{algorithmic}
\end{algorithm}

\begin{table*}[!t]
    \centering
    \caption{Reward terms. Each term contributes the product of its weight
    and its expression to its reward, and $\alpha$ is the scale in the last
    column. A hat denotes a target value: the reference value in the
    tracking terms and the commanded value in the task terms. Only the
    central task terms, which couple the task command $\bm c_t$ to the
    motion or the goal, are listed.}
    \label{tab:rewards}
    \footnotesize
    \setlength{\tabcolsep}{4pt}
    \begin{tabular}{@{}lllcc@{}}
        \toprule
        Reward & Term & Expression & Weight & Scale $\alpha$ \\
        \midrule
        Tracking & Root position
            & $\exp\!\big(-\alpha\|\Delta\bm p^{\,\mathrm{root}}\|_2^2\big)$
            & $0.5$ & $0.3^{-2}$ \\
        $r_t^{\mathrm{track}}$ ($\pi_E$) & Root orientation
            & $\exp\!\big(-\alpha\,\Delta\theta(\hat{\bm R}^{\,\mathrm{root}},
               \bm R^{\,\mathrm{root}})^2\big)$
            & $0.5$ & $0.4^{-2}$ \\
        & Body position
            & $\exp\!\big(-\alpha\langle\|\tilde{\bm p}^{\,b}-\bm p^{\,b}\|_2^2
               \rangle_{\mathcal{K}}\big)$
            & $1.0$ & $0.3^{-2}$ \\
        & Body orientation
            & $\exp\!\big(-\alpha\langle\Delta\theta(\tilde{\bm R}^{\,b},
               \bm R^{\,b})^2\rangle_{\mathcal{K}}\big)$
            & $1.0$ & $0.4^{-2}$ \\
        & Body lin.\ velocity
            & $\exp\!\big(-\alpha\langle\|\Delta\bm v^{\,b}\|_2^2
               \rangle_{\mathcal{K}}\big)$
            & $1.0$ & $1.0$ \\
        & Body ang.\ velocity
            & $\exp\!\big(-\alpha\langle\|\Delta\bm\omega^{\,b}\|_2^2
               \rangle_{\mathcal{K}}\big)$
            & $1.0$ & $3.14^{-2}$ \\
        \midrule
        Task $r_t^{\mathrm{task}}$ ($\pi_D$) & & & & \\
        \quad Flat terrain & Linear velocity
            & $\exp\!\big(-\alpha\|\hat{\bm v}_{xy}-\bm v_{xy}\|_2^2\big)$
            & $2.0$ & $5.0$ \\
        \quad and stairs & Yaw rate
            & $\exp\!\big(-\alpha(\hat{\omega}_z-\omega_z)^2\big)$
            & $2.5$ & $5.0$ \\
        \cmidrule(l{1em}){1-5}
        \quad Box climbing & Base height
            & $r_{\mathrm{base\mbox{-}height}}$
            & $1.0$ & \\
        & Foot on top surface
            & $r_{\mathrm{foot\mbox{-}surface}}$
            & $1.0$ & \\
        \cmidrule(l{1em}){1-5}
        \quad Recovery & Upright trunk
            & $\big(\bm n_{\mathrm{trunk}}^{\mathsf T}\bm e_z\big)^{+}$
            & $2.0$ & \\
        & Standing
            & $\mathbb{I}_{\mathrm{standing}}$
            & $1.0$ & \\
        \cmidrule(l{1em}){1-5}
        \quad Forward flip & Pitch tracking
            & $\exp\!\big(-\alpha(\theta_{\mathrm{pitch}}-
               \hat{\theta}_{\mathrm{pitch}}(t))^2\big)$
            & $5.0$ & $2.0$ \\
        & Feet landing
            & $\mathbb{I}_{\mathrm{feet\mbox{-}landing}}$
            & $2.5$ & \\
        \midrule
        Regularization & Undesired contact
            & $\sum_{b\in\mathcal{C}}\mathbb{I}[f^{\,b}>f_c]$
            & $-0.1$ & \\
        $r_t^{\mathrm{reg}}$ ($\pi_E$, $\pi_D$) & Action rate
            & $\|a_t-a_{t-1}\|_2^2$
            & $-0.1$ & \\
        & Action smoothness
            & $\|a_t-2a_{t-1}+a_{t-2}\|_2^2$
            & $-0.04$ & \\
        & Joint limit
            & $\sum_j\big[(q_j^{\min}-q_j)^+ + (q_j-q_j^{\max})^+\big]$
            & $-10.0$ & \\
        & Joint acceleration
            & $\|\ddot{\bm q}\|_2^2$
            & $-5\times10^{-8}$ & \\
        & Joint torque
            & $\|\bm\tau\|_2^2$
            & $-2\times10^{-5}$ & \\
        & Energy
            & $\sum_j|\tau_j\dot q_j|$
            & $-4\times10^{-4}$ & \\
        & Joint velocity
            & $\|\dot{\bm q}\|_2^2$
            & $-1\times10^{-3}$ & \\
        & Body sliding
            & $\sum_{b\in\mathcal{F}}\mathbb{I}[f^{\,b}>f_c]\,
               \|\bm v_{xy}^{\,b}\|_2$
            & $-0.1$ & \\
        \bottomrule
    \end{tabular}
\end{table*}
\subsection{Parallel-Policy Distribution Matching Algorithm}
\label{sec:parallel_training}

Each parallel simulation
environment is assigned to $\text{MDP}_E$ or $\text{MDP}_D$ at every
reset with a fixed probability.
In each iteration, the expert policy $\pi_E$ collects trajectories
$\tau_E^i$ of $T$ time steps in the environments assigned to
$\text{MDP}_E$ while tracking the reference motion clips $\mathcal{M}$, and the
downstream policy $\pi_D$ collects trajectories $\tau_D^i$ of the same
length in the environments assigned to $\text{MDP}_D$.
Algorithm~\ref{alg:PDM} denotes the number of trajectories of each policy
by $m$; under this assignment, the two numbers are equal in expectation.
At every
time step, the current discriminator evaluates the logits of the visited
states, from which the rewards $r_{E,t}$ and $r_{D,t}$ in
\eqref{eq:combined_rewards} are computed and recorded in the
corresponding trajectories. The states $s_{E,t}$ and $s_{D,t}$ visited in
these trajectories are stored in the discriminator buffers
$\mathcal{B}_E$ and $\mathcal{B}_D$, respectively. The discriminator is
then updated for $N_{\mathrm{disc}}$ steps, each of which minimizes the
objective in \eqref{eq:discriminator_loss} on mini-batches $b_E$ and
$b_D$ sampled from $\mathcal{B}_E$ and $\mathcal{B}_D$. Finally, each
policy and its critic, whose value function is $V_E$ or $V_D$, are
updated with proximal policy optimization
(PPO)~\cite{schulman2017proximal} using the trajectories and rewards of
that policy.

We trained both policies from scratch in the reported experiments.
PDM also supports initialization from a well-trained policy, which can
be held fixed to collect rollouts or fine-tuned during training.
This initialization can increase the magnitude of the discriminator
logits early in training, indicating stronger separation between the
two policies. We observed that the logits subsequently converged to the
equilibrium value.

\section{Implementation Details}
\label{sec:implementation_details}
This section describes the implementation used in the real-world
experiments in Section~\ref{sec:plane_real_world}.
Section~\ref{sec:reward_implementation} lists the rewards,
Section~\ref{sec:policy_inputs} describes the inputs to the policies,
critics, and discriminator, and Section~\ref{sec:training_schedule}
gives the training, domain-randomization, and deployment settings.
\subsection{Rewards}
\label{sec:reward_implementation}
Table~\ref{tab:rewards} lists the reward terms. The expert policy receives
the tracking reward $r_t^{\mathrm{track}}$, and the downstream policy
receives the task reward $r_t^{\mathrm{task}}$; in the real-world
experiments, both rewards also include the regularization reward
$r_t^{\mathrm{reg}}$.

\subsubsection{Tracking Reward}
\label{sec:tracking_reward}
Following the approaches in~\cite{peng2018deepmimic} and~
\cite{liao2025beyondmimic}, we compute $r_t^{\mathrm{track}}$ at each
control step by summing exponential kernels of the squared errors
between the robot and the time-synchronized reference motion clips
$\mathcal{M}$.
For the root (the base link) and each key body $b\in\mathcal{K}$,
$\bm p$, $\bm R$, $\bm v$, and $\bm\omega$ denote the position,
orientation, linear velocity, and angular velocity;
$\Delta\bm x=\hat{\bm x}-\bm x$; $\Delta\theta(\cdot,\cdot)$ is the angle
of the relative rotation between two orientations; and
$\langle\cdot\rangle_{\mathcal{K}}$ is the average over $\mathcal{K}$.
We compute the root and velocity terms in the world frame. For the
body-pose terms, we use the reference pose
$(\tilde{\bm p}^{\,b},\tilde{\bm R}^{\,b})$. We translate the reference
motion horizontally and rotate it about the vertical axis to align its
root with the robot root in horizontal position and heading, while
retaining the reference root height. The body-pose terms are therefore
unaffected by horizontal or heading drift of the root.

\subsubsection{Task Reward}
\label{sec:downstream_rewards}
The task reward specifies task completion without an online motion
reference. 
A pure tracking task uses the distribution-matching reward
and requires no additional task reward.
For the other tasks, we found that task rewards and termination
conditions helped the robot complete the task and mitigated mode
collapse. 

We use two reward settings.
For flat-terrain walking and blind stair locomotion, the task reward
adds auxiliary style-shaping terms, such as foot clearance and stance
duration, to the velocity-tracking terms. To compare the two methods
fairly in this general locomotion setting, we trained AMP with the same
terms.
For two-leg jumping, crawling, recovery, box climbing, and the forward
flip, the task reward contains only the central terms in
Table~\ref{tab:rewards} and no style-shaping terms. These terms reward
the task goal but do not specify how the robot reaches it.
Combined with the kinematically retargeted reference motions, these minimal rewards create
an extreme setting in which the motion is guided mainly by the
distribution-matching reward.
More elaborate task rewards may improve performance on these tasks; we
keep them minimal to isolate the effect of distribution matching.
Table~\ref{tab:rewards} lists the central terms of all tasks and omits
the auxiliary shaping terms.

Table~\ref{tab:task_commands} summarizes the task command $\bm c_t$ for
each task.
For walking, two-leg jumping, crawling, and blind stair locomotion,
$\bm c_t=(\hat{\bm v}_{xy},\hat{\omega}_z)$ is the commanded horizontal velocity and yaw rate,
which are tracked in a yaw-aligned frame.
For box climbing,
$\bm c_t$ is a completion signal indicating whether the climb has finished,
and $r_{\mathrm{foot\mbox{-}surface}}$ rewards the feet for reaching the top surface of the box.
For recovery,
$\bm c_t$ is a completion signal indicating whether recovery has finished.
Here, $\bm n_{\mathrm{trunk}}$ is the upward normal of the trunk,
$\bm e_z$ is the upward vertical direction,
$(\cdot)^+$ is the positive part,
and $\mathbb{I}_{\mathrm{standing}}$ indicates a stable standing state;
a sustained-standing bonus rewards task completion rather than a
momentary upright pose.
For the forward flip,
$\bm c_t$ contains the commanded pitch rate and a flag that indicates whether the flip has been completed.
The target $\hat{\theta}_{\mathrm{pitch}}(t)$ of the trunk pitch $\theta_{\mathrm{pitch}}$ progresses from $0$ to $2\pi$ over the desired rotation interval,
and $\mathbb{I}_{\mathrm{feet\mbox{-}landing}}$ indicates a feet-first landing.

\subsubsection{Regularization Reward}
\label{sec:regularization_rewards}
The regularization reward $r_t^{\mathrm{reg}}$ includes penalties on
contacts, actions, and joint-level quantities.
In Table~\ref{tab:rewards},
$a_t$ is the action;
$\bm q$, $\dot{\bm q}$, $\ddot{\bm q}$, and $\bm\tau$ are the joint positions, velocities, accelerations, and torques;
and $q_j^{\min}$ and $q_j^{\max}$ are the soft limits of joint $j$.
The quantities $f^{\,b}$ and $\bm v_{xy}^{\,b}$ are the contact-force magnitude and horizontal velocity of body $b$,
and $f_c$ is a contact threshold.
The set $\mathcal{F}$ contains the feet,
$\mathcal{C}$ contains all bodies except the feet and the elbow-yaw links.
Table~\ref{tab:rewards} reports the default weights, which the expert
and downstream policies share.
Box climbing and the forward flip use weights of $-0.5$ and $-0.08$ for
the action-rate and action-smoothness terms, and box climbing also uses
weights of $-1\times10^{-7}$, $-1\times10^{-5}$, and $-2\times10^{-3}$
for the joint-acceleration, joint-torque, and joint-velocity terms.
The active terms also vary across tasks: box climbing omits the energy
term, the recovery task omits the contact penalty, and the recovery and
crawling tasks apply the sliding penalty to all bodies.

\subsection{Reference Motion Preparation and Usage}
\label{sec:reference_processing}
Demonstration quality affects imitation learning performance,
particularly in adversarial approaches. Because adversarial rewards are
difficult to interpret, poor task learning can be hard to attribute to
either the motion data or the choice of algorithm hyperparameters.
More sophisticated retargeting methods can improve data quality but
often require substantial expertise and specialized optimization.
Motions optimized for a nominal robot model may also need further
adaptation to the randomized dynamics used during policy training.
Our tracking expert learns to execute reference motions under domain
randomization. This simplifies retargeting and produces trajectories
that account for robot dynamics and contact interactions.

We use basic kinematic retargeting to prepare the reference motions.
The tracking policy compensates for retargeting inaccuracies and
accounts for dynamics and contact constraints.
We map the original motion-capture data directly to the robot's
configuration space using the kinematic retargeting procedure described
in~\cite{araujo2025retargeting}.
The motions used in the real-world experiments came primarily from the
LAFAN1 dataset~\cite{harvey2020robust}. We trimmed each sequence to the
segment needed for the corresponding downstream task and collected the
remaining clips with a motion-capture device.

During training, expert agents can start from any reference state,
whereas downstream agents can start from either a reference state or the
default initial state. We apply early termination to expert agents to
facilitate motion tracking. Downstream agents use the termination
conditions of their respective tasks.

\begin{table}[t]
    \centering
    \caption{Network inputs. A check mark indicates that the network
    receives the term, and an open circle $\circ$ indicates an optional or
    task-dependent input.}
    \label{tab:network_inputs}
    \footnotesize
    \setlength{\tabcolsep}{3pt}
    \begin{tabular}{@{}lccccc@{}}
        \toprule
        & \multicolumn{2}{c}{Expert} & \multicolumn{2}{c}{Downstream} & \\
        \cmidrule(lr){2-3}\cmidrule(lr){4-5}
        Input & Actor & Critic & Actor & Critic & Disc. \\
        \midrule
        \multicolumn{6}{@{}l}{\emph{Proprioception} $\bm o_t^{\mathrm{prop}}$} \\
        \quad Root angular velocity
            & $\checkmark$ & $\checkmark$ & $\checkmark$ & $\checkmark$ & $\checkmark$ \\
        \quad Gravity direction
            & $\checkmark$ & $\checkmark$ & $\checkmark$ & $\checkmark$ & $\checkmark$ \\
        \quad Joint positions
            & $\checkmark$ & $\checkmark$ & $\checkmark$ & $\checkmark$ & $\checkmark$ \\
        \quad Joint velocities
            & $\checkmark$ & $\checkmark$ & $\checkmark$ & $\checkmark$ & $\checkmark$ \\
        \quad Previous action
            & $\checkmark$ & $\checkmark$ & $\checkmark$ & $\checkmark$ & $\circ$ \\
        \midrule
        \multicolumn{6}{@{}l}{\emph{Reference} $\bm o_t^{\mathrm{ref}}$} \\
        \quad Joint positions
            & $\checkmark$ & $\checkmark$ & & & \\
        \quad Joint velocities
            & $\checkmark$ & $\checkmark$ & & & \\
        \quad Delta root position
            & $\checkmark$ & $\checkmark$ & & & \\
        \quad Delta root orientation
            & $\checkmark$ & $\checkmark$ & & & \\
        \quad Key-body poses
            & $\checkmark$ & $\checkmark$ & & & \\
        \midrule
        \multicolumn{6}{@{}l}{\emph{Privileged state} $\bm o_t^{\mathrm{priv}}$
            \emph{and other inputs}} \\
        \quad Root height
            & $\circ$ & $\checkmark$ & & $\checkmark$ & $\checkmark$ \\
        \quad Root linear velocity
            & $\checkmark$ & $\checkmark$ & & $\checkmark$ & $\checkmark$ \\
        \quad Key-body poses
            & $\checkmark$ & $\checkmark$ & & $\checkmark$ & $\checkmark$ \\
        \quad Key-body velocities
            & $\circ$ & $\checkmark$ & & $\checkmark$ & $\checkmark$ \\
        \quad Height scan $\bm h_t$
            & $\circ$ & $\circ$ & & $\circ$ & \\
        \quad Task command $\bm c_t$
            & & & $\checkmark$ & $\checkmark$ & \\
        \midrule
        History length (steps)
            & $1$ & $1$ & $10$ & $10$ & $3$ \\
        \bottomrule
    \end{tabular}
\end{table}

\begin{table}[t]
    \centering
    \caption{Task command $\bm c_t$ of the downstream policy.}
    \label{tab:task_commands}
    \footnotesize
    \begin{tabular}{@{}llc@{}}
        \toprule
        Task & Command $\bm c_t$ & Dim. \\
        \midrule
        Flat-terrain walking
            & Velocity and yaw rate
              $(\hat{\bm v}_{xy},\hat{\omega}_z)$ & $3$ \\
        Two-leg jumping
            & Velocity and yaw rate
              $(\hat{\bm v}_{xy},\hat{\omega}_z)$ & $3$ \\
        Crawling
            & Velocity and yaw rate
              $(\hat{\bm v}_{xy},\hat{\omega}_z)$ & $3$ \\
        Blind stair locomotion
            & Velocity and yaw rate
              $(\hat{\bm v}_{xy},\hat{\omega}_z)$ & $3$ \\
        \midrule
        Recovery & Completion signal & $1$ \\
        Box climbing & Completion signal & $1$ \\
        \midrule
        Forward flip & Pitch rate and completion signal & $2$ \\
        \bottomrule
    \end{tabular}
\end{table}
\subsection{Network Architectures and Inputs}
\label{sec:policy_inputs}
\subsubsection{Network Architectures}
The expert and downstream actors and critics, as well as the
discriminator, are multilayer perceptrons (MLPs) with hidden layers of
512, 256, and 128 units and exponential linear unit (ELU) activations.
We normalize discriminator, all actor and
critic inputs using running estimates of their mean and standard
deviation.

\subsubsection{Actor and Critic Inputs}
Table~\ref{tab:network_inputs} lists the actor and critic inputs.
Key-body poses and velocities are expressed in the root frame,
reference root quantities are expressed relative to the robot root, and
root height is expressed in the world frame.
Both branches use asymmetric actor--critic training, in which each critic
also receives privileged simulation quantities. The inputs comprise
proprioception $\bm o_t^{\mathrm{prop}}$ measured on the robot;
reference observations $\bm o_t^{\mathrm{ref}}$ taken from the frame of
$\mathcal{M}$ at the current motion time; privileged state
$\bm o_t^{\mathrm{priv}}$, which contains body-state quantities that are
not measured on the robot; a terrain height scan $\bm h_t$, which is
used only in blind stair locomotion and box climbing; and the task
command $\bm c_t$. The actor
columns of Table~\ref{tab:network_inputs} define the policy inputs
$\bm o_E$ and $\bm o_D$. Because the expert actor is used only during
training, it also receives part of $\bm o_t^{\mathrm{priv}}$. The
downstream actor receives only quantities available on the robot.
It does not observe $\mathcal{M}$, expert samples, or a reference frame
or phase. It also receives no terrain input in blind stair locomotion
and no box height in box climbing.
Exteroceptive observations may improve performance on terrain-interaction
tasks, but incorporating them requires additional engineering beyond
the scope of this paper.

\subsubsection{Discriminator Input}
Previous adversarial imitation learning
frameworks~\cite{peng2021amp} rely on features available from reference
motion data, which may contain noise from motion capture or retargeting.
These inaccuracies can make some features easy to distinguish and cause
the discriminator to rely excessively on them. PDM allows the
discriminator to use a flexible set of features with ground-truth values
from the simulator.

We construct the feature vector $\Phi(s_t)$ identically for expert and
downstream samples. It comprises root height, root linear velocity,
root angular velocity, gravity direction, joint positions, joint
velocities, and key-body poses and velocities, as listed in
Table~\ref{tab:network_inputs}.
The discriminator receives a history of $H$ rollout steps, with $H=3$,
$(\Phi(s_{t-H+1}),\ldots,\Phi(s_t))$.
Actions can also be included in this history, although they had a
negligible effect on performance in our experiments, possibly because
the state history already captures some of their effects.
The discriminator can also incorporate per-body contact signals and
terrain information; we leave their effects for future study.

\begin{table}[t]
    \centering
    \caption{PDM training hyperparameters. GAE denotes generalized
    advantage estimation.}
    \label{tab:pdm_hyperparams}
    \begin{tabular}{lcc}
        \toprule
        Parameter & Expert policy & Downstream policy \\
        \midrule
        Steps per rollout            & $24$ & $24$ \\
        PPO mini-epochs              & $5$ & $5$ \\
        PPO mini-batches             & $4$ & $4$ \\
        Discount factor $\gamma$     & $0.99$ & $0.99$ \\
        GAE parameter $\lambda$      & $0.95$ & $0.95$ \\
        PPO clip range $\epsilon$    & $0.2$ & $0.2$ \\
        Value-loss coefficient       & $1.0$ & $1.0$ \\
        Entropy coefficient          & $0.01$ & $0.01$ \\
        Learning rate                & $3\times10^{-5}$ & $1\times10^{-5}$ \\
        Gradient clipping            & $1.0$ & $1.0$ \\
        KL weight $\beta$            & $0.0$ & $1.0$ \\
        \midrule
         & \multicolumn{2}{c}{Discriminator} \\
        \midrule
        Learning rate                & \multicolumn{2}{c}{$2\times10^{-5}$} \\
        Weight decay                 & \multicolumn{2}{c}{$1\times10^{-5}$} \\
        Mini-epochs                  & \multicolumn{2}{c}{$2$} \\
        Mini-batches                 & \multicolumn{2}{c}{$2$} \\
        Replay buffer size           & \multicolumn{2}{c}{$200{,}000$} \\
        New samples                  & \multicolumn{2}{c}{$20{,}000$} \\
        \bottomrule
    \end{tabular}
\end{table}
\subsection{Training and Deployment Settings}
\label{sec:training_schedule}
\label{sec:real_world_settings}

We trained the policies for the real-world experiments with RSL-RL~\cite{schwarke2025rslrl} and
IsaacLab~\cite{mittal2025isaaclab}, using the
hyperparameters in Table~\ref{tab:pdm_hyperparams}. A training run takes
approximately 20~hours on a single NVIDIA RTX 4090 GPU, with some
variation across skills. We simulate 8{,}192 parallel environments.
At every reset, we reassign each environment to $\text{MDP}_E$ or
$\text{MDP}_D$ with a fixed probability of $0.5$. This gives a $1{:}1$
allocation between the branches, with 4{,}096 environments per branch in
expectation, rather than a fixed set assigned to each branch.
Because some domain-randomization terms are
sampled only once at startup, this reassignment exposes both branches to
the full randomization ranges. With $\beta_E>0$, distribution matching
also draws the expert policy toward the downstream distribution, so
both policies match the other's distribution while pursuing their own
rewards. With $\beta_E=0$, as in Table~\ref{tab:pdm_hyperparams}, the
expert is a pure tracking policy unaffected by the downstream branch,
and information flows in one direction.

\begin{table}[b]
    \centering
    \caption{Domain-randomization terms. CoM denotes center of mass.
    Parameters are sampled uniformly from the listed ranges.}
    \label{tab:domain_randomization}
    \begin{tabular}{ll}
        \toprule
        Term & Range \\
        \midrule
        \multicolumn{2}{l}{\emph{Sampled once per environment at startup}} \\
        Static friction coefficient
            & $[0.3,1.6]$ \\
        Dynamic friction coefficient
            & $[0.3,1.2]$ \\
        Restitution coefficient
            & $[0,0.5]$ \\
        Base CoM offset
            & $\pm(0.025,0.05,0.05)$~m \\
        Default joint-position offset
            & $\pm0.01$~rad \\
        Joint stiffness scale
            & $[0.9,1.1]$ \\
        Joint damping scale
            & $[0.75,1.5]$ \\
        \midrule
        \multicolumn{2}{l}{\emph{Applied at random intervals of $1$--$3$~s}} \\
        Root linear-velocity push
            & $\pm(0.5,0.5,0.2)$~m/s \\
        Root angular-velocity push
            & $\pm(0.52,0.52,0.78)$~rad/s \\
        \midrule
        \multicolumn{2}{l}{\emph{Resampled after environment resets}} \\
        Action delay
            & $[0.02,0.08]$~s \\
        \bottomrule
    \end{tabular}
\end{table}

We conducted all real-world experiments on the PM01 humanoid robot,
which has a mass of approximately $42$~kg and a standing height of
approximately $1.5$~m. We used 23 joints: five in each arm, six in each
leg, and one in the torso.
We trained both policies with the domain-randomization settings in
Table~\ref{tab:domain_randomization}. To improve generalization in blind
stair locomotion and box climbing, we also randomized the terrain
geometry. For box climbing, we randomized the box height within
\([0.4,0.8]\)~m, while the reference motion corresponded to a fixed box
height of \(0.4\)~m. For blind stair locomotion, we sampled stair heights
uniformly from \([0,0.25]\)~m.
To further reduce the sim-to-real gap, we added observation noise only
to the downstream actor inputs. The expert actor, discriminator, and
both critics received noise-free observations.
We set the downstream actor's ankle-joint velocity inputs to zero to
mitigate the mismatch between the simulated ankle model and the
parallel-ankle mechanism on hardware.
After training, we deployed only the downstream policy on hardware,
using proprioceptive observations and task commands.

\begin{figure*}[t]
    \centering
    \includegraphics[trim=0bp 3.5bp 0bp 6bp, clip]{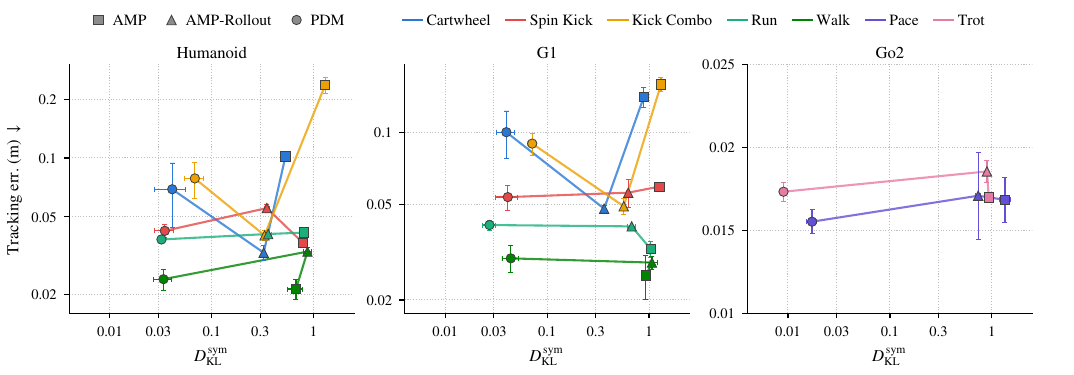} 
    \caption{Tracking position error versus the symmetric-KL logit-gap
    diagnostic for AMP, AMP-Rollout, and
    PDM (Ours) on 12 motion clips across three platforms. The horizontal
    axes use logarithmic scales.}
    \label{fig:trade_off}
\end{figure*}
\section{Simulation Results}
\label{sec:simulation_results}
We used a setup different from that of the real-world experiments
because our AMP baseline learned some downstream tasks poorly under the
extreme settings.
To compare the methods in this section fairly, we trained all of them
with MimicKit~\cite{MimicKitPeng2025}, which provides a well-tuned AMP
implementation.
We reimplemented the PDM method described in Section~\ref{sec:method}
in MimicKit.
The downstream policy $\pi_D$ and the discriminator used the parameters
of the AMP baseline, and the tracking policy $\pi_E$ used the DeepMimic
pipeline.
To focus on adversarial training, we exclude latent-conditioned and
diffusion methods and compare PDM with two adversarial baselines:
AMP~\cite{peng2021amp} and AMP-Rollout. The three methods are as follows.
\begin{enumerate}
\item \textbf{AMP:}
AMP uses the standard adversarial motion imitation setting~\cite{peng2021amp}. Its expert
dataset is the kinematically retargeted motion clip, fixed before policy
optimization and unchanged throughout training

\item \textbf{AMP-Rollout:}
We introduce an AMP variant that adopts the \emph{collect-then-imitate}
structure of GAIL~\cite{ho2016generative} but not its original
implementation.
We first trained a tracking policy with DeepMimic~\cite{peng2018deepmimic}
and then evaluated the policy to generate the trajectory that the robot
actually executed. This trajectory served as the entire expert dataset
and remained fixed throughout adversarial training.

\item \textbf{PDM(Ours):}
In PDM, the tracking policy $\pi_E$ and the downstream task policy $\pi_D$
are trained concurrently from scratch. As $\pi_E$ moves closer to the
reference motion, part of its rollout replaces the expert buffer, and the
adversarial distribution-matching reward guides $\pi_D$ to match the
evolving expert distribution.
\end{enumerate}

The comparison examines two factors.
First, AMP uses kinematically retargeted expert data for the
discriminator, whereas AMP-Rollout and PDM use expert data executed by
the robot.
Second, AMP-Rollout records the expert motion with deterministic action
sampling, whereas PDM generates expert rollouts online with stochastic
action sampling during training.
To examine how these factors affect adversarial motion imitation, we
evaluated single-clip motion imitation on 12 motion clips across three
platforms and ran each training configuration with three random seeds.
For evaluation, we used a fixed random seed and started each rollout from
the first frame of the reference motion.
To keep the discriminator rewards consistent across methods, all
adversarial methods used the logarithmic reward mapping in
Fig.~\ref{fig:discrinmiantor_reward} in the tracking, logit, and
mechanism experiments.

\subsection{Comparison study}
We assess tracking performance by measuring the similarity between the
robot's motion and the reference
motion~\cite{peng2018deepmimic,peng2021amp}.
The tracking position error $e_t^{\mathrm{track}}$ combines the global
root-position error with the root-relative joint-position errors:
\begin{equation}
\begin{aligned}
e_t^{\mathrm{track}}
&= \frac{1}{N_{\mathrm{J}}+1}
    \Biggl(
    \bigl\|
        \hat{\bm{p}}_t^{\,\mathrm{root}}-\bm{p}_t^{\,\mathrm{root}}
    \bigr\|_2\\
&\qquad {}+
    \sum_{j\in\mathrm{J}}
    \Bigl\|
        \bigl(\hat{\bm{p}}_t^{\,j}-\hat{\bm{p}}_t^{\,\mathrm{root}}\bigr)
        -
        \bigl(\bm{p}_t^{\,j}-\bm{p}_t^{\,\mathrm{root}}\bigr)
    \Bigr\|_2 
    \Biggr),
\end{aligned}
\label{eq:pos_error}
\end{equation}
where $\bm p_t^{\,j}$ and $\hat{\bm p}_t^{\,j}$ denote the three-dimensional Cartesian
positions of joint $j$ in the simulated robot and reference motion,
respectively, and $N_{\mathrm{J}}$ is the number of joints.
By including both global root-position error and root-relative joint
errors, $e_t^{\mathrm{track}}$ captures deviations of the root and errors
in body configuration. We average this error over all time steps of the
evaluation episodes.
We also report a diagnostic of how readily each discriminator
distinguishes its expert and policy samples.  We define this
metric as
\begin{equation}
D^\text{KL}_\text{sym} =
\begin{cases}
\mathbb{E}_{\rho_E}[z^*]
- \mathbb{E}_{\rho_D}[z^*],
& \text{PDM}, \\
\mathbb{E}_{\mathcal M}[z^*]
- \mathbb{E}_{\rho_D}[z^*],
& \text{AMP/AMP-Rollout}.
\end{cases}
\label{eq:logit_gap}
\end{equation}
Because each policy is trained against its own discriminator,
$D^\text{KL}_\text{sym}$ serves only as a relaxed comparison across
methods. It indicates overall trends rather than a strict ranking.
\begin{figure*}[t]
    \centering
    \includegraphics[trim=0bp 2.6bp 0bp 6bp, clip]{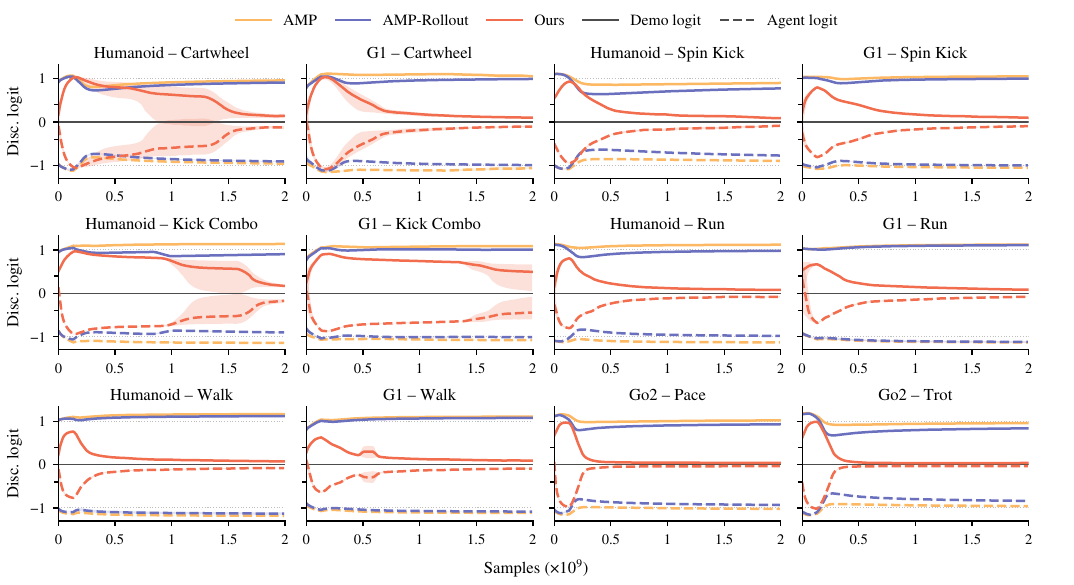} 
    \caption{Demo (solid) and agent (dashed) discriminator logits of AMP, AMP-Rollout, and PDM during training on each motion clip.}
    \label{fig:training_logits}
\end{figure*}
Figure~\ref{fig:trade_off} compares AMP, AMP-Rollout, and PDM after about
2 billion training samples.
PDM achieved lower tracking error than AMP on 7 of the 12 clips, and
AMP-Rollout achieved lower tracking error than AMP on 6 of the 12 clips.
Both methods reduced the error most on the Cartwheel and Kick Combo clips
of Humanoid and G1, where AMP-Rollout achieved the lowest error.
On the Run and Walk clips of Humanoid and G1, AMP achieved tracking error
comparable to or lower than that of AMP-Rollout.
AMP-Rollout yielded a smaller $D^\text{KL}_\text{sym}$ than AMP on 10 of
the 12 clips, and PDM yielded the smallest $D^\text{KL}_\text{sym}$ on all
12 clips.

For pure tracking tasks, these results suggest that reference-trajectory
quality is a primary factor in tracking accuracy.
Physically executable reference motions provided the largest benefits on
the Cartwheel and Kick Combo clips and limited benefits on running and
walking, where the retargeted trajectories were already within the robot's
capabilities.
On the Cartwheel and Kick Combo clips, AMP-Rollout was the most accurate
but requires a separate tracking-policy training stage, whereas PDM avoids
this stage and still reduced the tracking error relative to AMP.
The remaining gap between PDM and AMP-Rollout may arise because the
training budget was insufficient for the tracking policy to converge and
because the stochastic rollout buffer can suppress tracking performance.
Longer training may further reduce the tracking error at the cost of
additional training time.

Under the relaxed $D^\text{KL}_\text{sym}$ comparison, the smaller value
of AMP-Rollout relative to AMP suggests that improving reference-motion
quality makes expert and policy samples harder to distinguish.
PDM obtained the smallest value, which is consistent with the logits
approaching equilibrium as both policies generate evolving samples for
discriminator training and suggests stronger distribution matching between
the expert and downstream policies.
This stronger matching may offer greater benefits for downstream tasks,
where the policy must preserve the desired motion style while adapting
to task requirements.

\begin{figure}[!t]
    \centering
    \includegraphics[trim=0bp 5bp 0bp 20bp, clip]{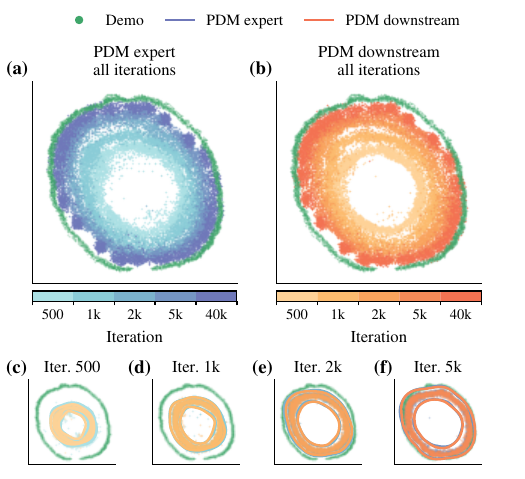}
    \caption{Uniform manifold approximation and projection (UMAP) of the state transitions of the PDM expert and downstream policies during training on the G1--Run clip. (a), (b) Expert and downstream samples at iterations 500, 1k, 2k, 5k, and 40k, together with the reference motion (Demo). (c)--(f) Samples of both policies and of the reference motion at iterations 500, 1k, 2k, and 5k.}
    \label{fig:distribution}
    \vspace{-2em}
\end{figure}
\begin{figure*}[t]
    \centering
    \includegraphics[trim=0mm 0mm 0mm 0mm, clip]{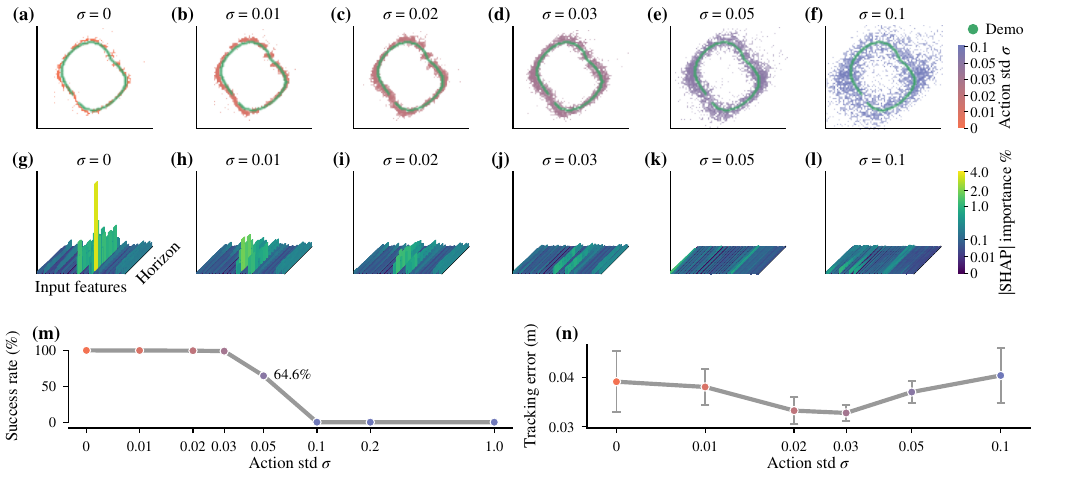}
    \caption{Effect of the expert action standard deviation $\sigma$ on the G1--Run clip. (a)--(f) PCA of the expert state occupancy and the reference motion (Demo). (g)--(l) Absolute SHAP importance (\%) of the discriminator input features. (m) Expert success rate. (n) Final downstream tracking error.}
    \label{fig:freeze_vs_normal}
    \vspace{-2em}
\end{figure*}
\subsection{Logit and Distribution Training Dynamics}
The most distinct difference concerns discriminator-logit convergence.
The discriminator can become overconfident and saturate early in training, making its rewards less informative~\cite{li2023learning,li2025featuregan}.
Retargeting artifacts and domain randomization can aggravate this
problem.
The discriminator often relies on features other than those used for
frame-aligned tracking. This allows it to replace manually designed
rewards, but noisy features can cause early saturation.
As the policy improves, the logits decrease. If the discriminator
becomes too strong and dominates training, policy samples receive small
rewards that can be obscured by noise, reducing learning efficiency and
performance.
These dynamics make it difficult to choose hyperparameters that work
across tasks. Pure tracking is a special case in which the logits can
reach equilibrium. With an additional task reward, the logits can
diverge as the agent omits some motions, leaving a larger reverse-KL gap.

We visualize the state-transition distributions during training to examine how PDM moves both policies toward the reference motion.
We collected state transitions $(s,s')$ from the expert and downstream
policies on the G1--Run clip at different training iterations, together
with reference-motion transitions (Demo), as shown in Fig.~\ref{fig:distribution}.
The policies initially occupy different regions of the state space. 
Their distributions move closer during training, 
while tracking rewards guide the expert to
produce transitions near the reference trajectories.
Using only the matching reward, the downstream policy matches the
evolving expert distribution and eventually reproduces the reference trajectory. 
The AMP baselines instead match the reference distribution directly and suffer from early saturation.

\begin{figure}[!t]
    \centering
    \includegraphics[trim=0bp 2bp 0bp 2bp, clip]{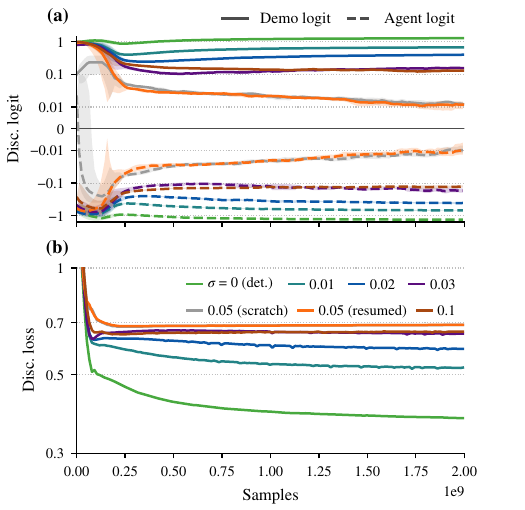}
    \caption{(a) Demo (solid) and agent (dashed) discriminator logits and (b) discriminator loss of PDM for different expert action standard deviations $\sigma$. scratch: expert trained from scratch.}
    \label{fig:resume_logits}
    \vspace{-2em}
\end{figure}
\subsection{Stochastic Action Noise Ablation}
The stochastic rollout buffer is a key design choice in PDM and may be the main factor that determines logit convergence.
To isolate the effect of action sampling in the expert rollouts, we loaded the expert from the same well-trained 20k-iteration checkpoint and held it fixed.
The loaded expert received no further PPO updates and only generated rollouts for downstream-policy training.
The fixed expert sampled actions with standard deviation $\sigma\in\{0, 0.01, 0.02, 0.03, 0.05, 0.1\}$, where $\sigma=0$ denotes deterministic sampling.
In addition, we trained an expert from scratch with $\sigma=0.05$ (labeled ``scratch'' in Fig.~\ref{fig:resume_logits}) and updated it concurrently with the downstream policy using PPO.
This run follows the standard PDM setting used in the tracking experiments above and also serves as a reference for the fixed experts.
The downstream policy used a fixed action standard deviation of 0.05 in all runs.

The logit and loss curves in Fig.~\ref{fig:resume_logits} differ between deterministic and stochastic sampling.
Under deterministic sampling, the expert and downstream logits initially approach equilibrium but then rapidly diverge before stabilizing, a pattern similar to that of AMP.
As $\sigma$ increases from 0.01 to the matched value of 0.05, the expert and downstream logits stabilize progressively closer to zero, while the discriminator loss reaches higher plateaus.
Both $\sigma=0.05$ cases yield similar small nonzero logits and losses near 0.69, regardless of expert initialization.
At $\sigma=0.1$, the expert logit and the loss settle near the levels observed at $\sigma=0.03$.

These logit trends coincide with changes in the expert state occupancy and in the absolute SHAP importance of the discriminator input features.
The projections onto the first two principal components in Fig.~\ref{fig:freeze_vs_normal}(a)--(f) show increasingly broad state-space coverage around the reference motion as $\sigma$ increases.
However, the expert success rate in Fig.~\ref{fig:freeze_vs_normal}(m) decreases as $\sigma$ increases and reaches zero at $\sigma=0.1$.
The absolute SHAP importance in Fig.~\ref{fig:freeze_vs_normal}(g)--(l) shows sharp peaks under deterministic sampling.
As $\sigma$ increases to 0.05, the importance shifts to lower values that are more evenly distributed across features and the horizon.
At $\sigma=0.1$, the importance becomes less evenly distributed than at $\sigma=0.05$.
The final downstream tracking error in Fig.~\ref{fig:freeze_vs_normal}(n) decreases as $\sigma$ increases from 0 to 0.03, where it reaches its lowest value, and increases at larger $\sigma$.

Together, these observations associate reduced logit separation with broader sample coverage and less concentrated feature importance.
We attribute this difference to stochastic sampling, which increases both the diversity of expert trajectories and the overlap between the expert and downstream distributions.
The broader distribution of expert samples over the state space reduces overfitting of the discriminator to the narrow set of states visited by the expert.
This reduction improves matching performance, whereas excessive action noise degrades the fidelity of the expert motion and limits this improvement.
This finding suggests that discriminator matching performance depends not only on the fidelity of the expert motion but also on how expert samples are generated.
In particular, even a well-trained expert can limit matching performance if its samples are generated either deterministically or with large action noise.

\subsection{Mixing Ratio Ablation}
To study how the expert's distribution-matching reward affects
training, we mix the tracking and distribution-matching rewards using
the ratio $\beta_E$:
\begin{equation}
r_{E} = (1-\beta_E) r_\mathrm{track} + \beta_E f_E,
\end{equation}
and report the raw values of $r_\mathrm{track}$ and $f_E$, without weighting.
When $\beta_E \rightarrow 0$, the KL regularizer no longer acts on the
expert branch. This limit coincides with the setting $\beta_E=0$ in
\eqref{eq:expert_combined_reward} used for the baseline comparison.
When $\beta_E \rightarrow 1$, the expert receives only the
distribution-matching reward.
\begin{figure}[h]
    \centering
    \includegraphics[trim=0bp 2.4bp 0bp 2.5bp, clip,width=\linewidth]{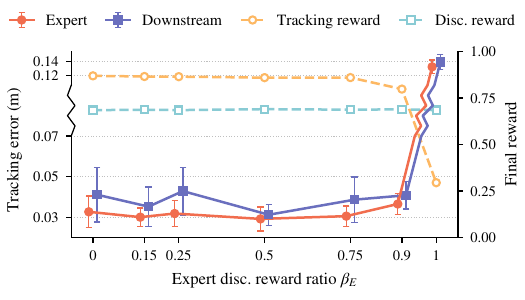}\\[2pt]
    \caption{Effect of the mixing ratio $\beta_E$ on the tracking errors of the expert and downstream policies (solid lines, left axis) and on the final tracking and discriminator rewards (dashed lines, right axis).}
    \label{fig:tracking_ratio}
\end{figure}

We tested $\beta_E\in\{0.15, 0.25, 0.5, 0.75, 0.9, 1\}$ on the G1--Run
clip. For each ratio, Fig.~\ref{fig:tracking_ratio} reports the
tracking errors of the expert and downstream policies and the final
values of $r_{track}$ and $f_E$.
For $0.15\leq\beta_E\leq 0.75$, the tracking reward remains between 0.86
and 0.87, whereas the discriminator reward remains at approximately 0.69
for every tested $\beta_E$; therefore, a larger $\beta_E$ reduces
tracking accuracy without increasing the discriminator reward.
For $\beta_E\leq 0.9$, the downstream tracking error remains between
0.031 and 0.043~m.
As $\beta_E$ approaches 1, however, both errors increase sharply, and at
$\beta_E=1$, both policies fail to complete the task.
These results indicate that the tracking accuracy of the downstream
policy depends on that of the expert: when the expert loses tracking
accuracy, the downstream policy loses it as well.
\begin{figure}[ht]
    \centering
    \includegraphics[width=\linewidth]{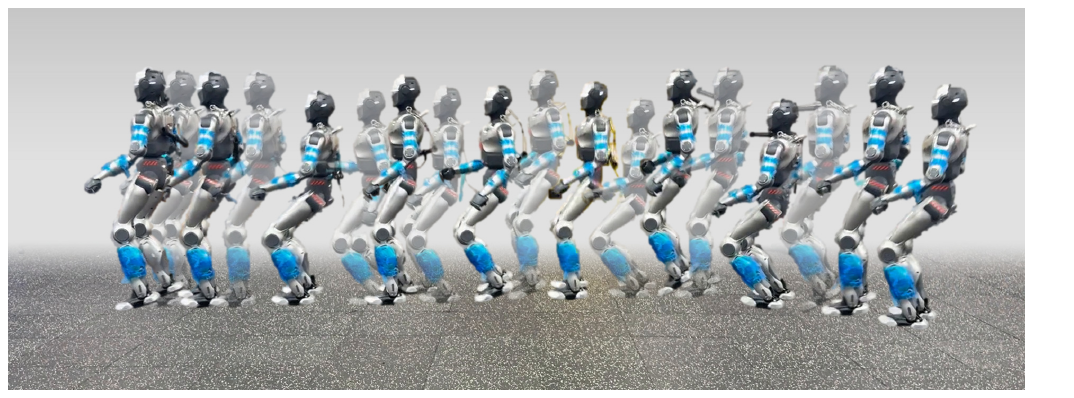}\\[2pt]
    \includegraphics[width=\linewidth]{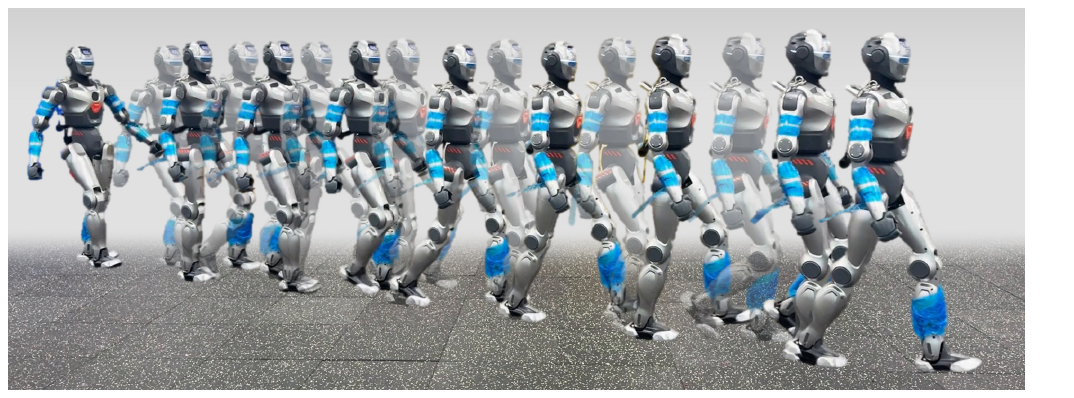}\\[2pt]
    \includegraphics[width=\linewidth]{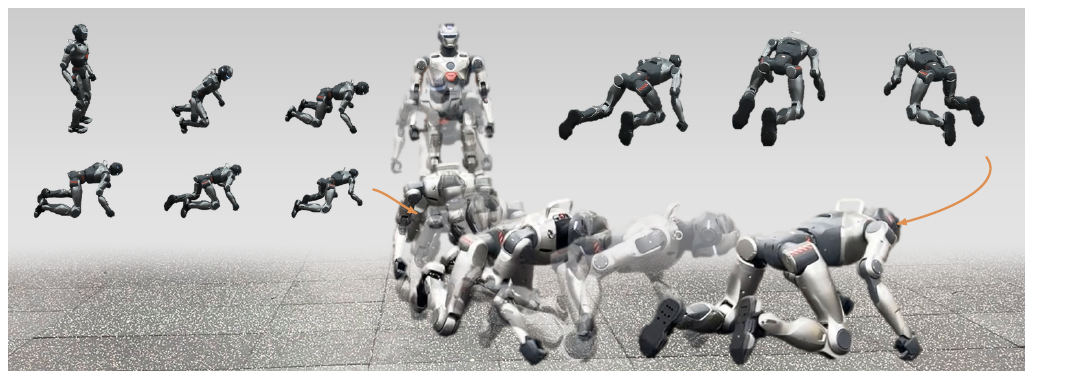}
    \caption{Real-world flat-terrain skills of the deployed downstream policies on the PM01 humanoid robot: two-leg jumping (top), walking (middle), and crawling (bottom).}
    \label{fig:plane}
    \vspace{-1.5em}
\end{figure}
\begin{figure*}[t]
    \centering
    \subfloat[]{
        \includegraphics[width=0.5\linewidth]{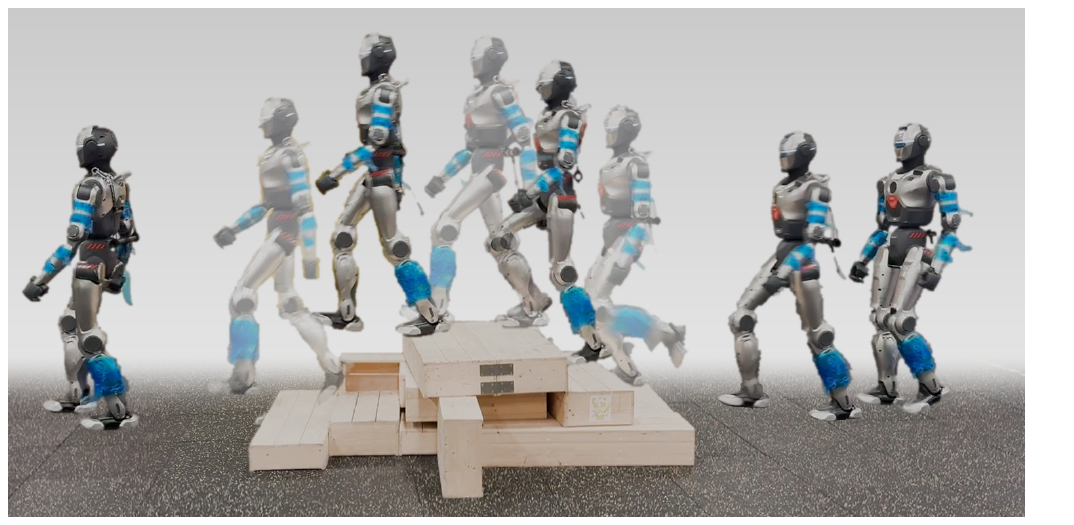}
    }
    \subfloat[]{
        \hspace{-1em}
        \includegraphics[width=0.5\linewidth]{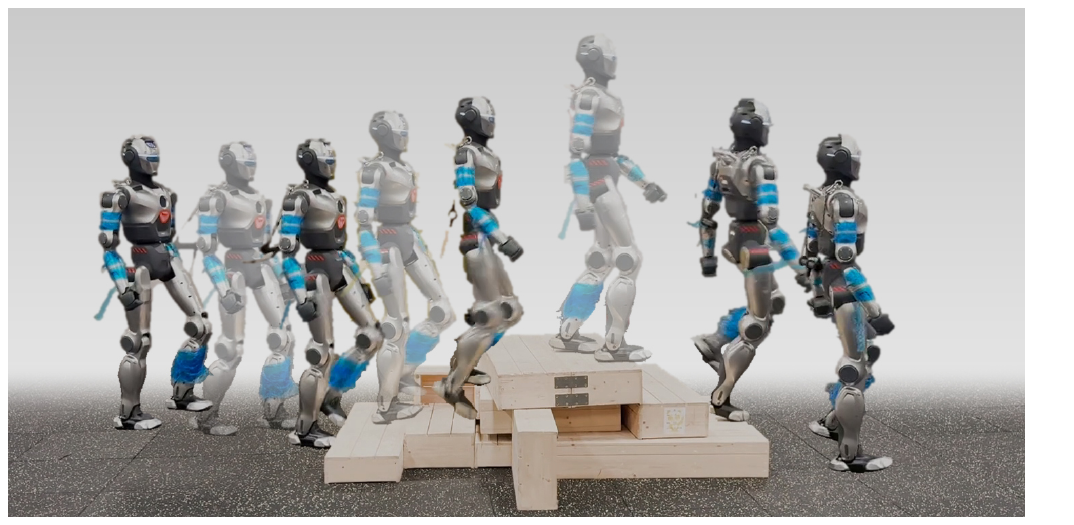}
        \label{fig:stair_right}
    }\\
    \subfloat[Low platform]{
        \includegraphics{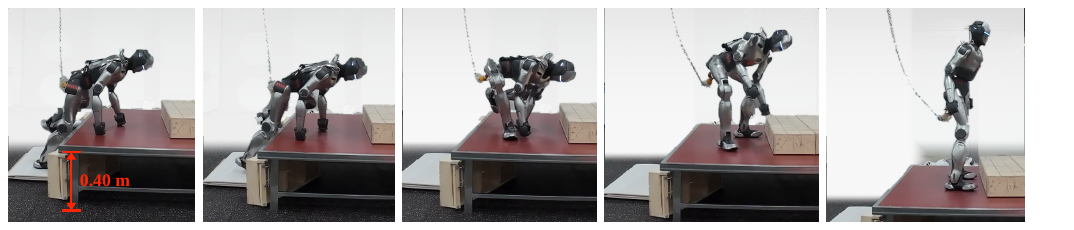}
        \label{fig:climb_low}
    }\\
    \subfloat[High platform]{
        \includegraphics{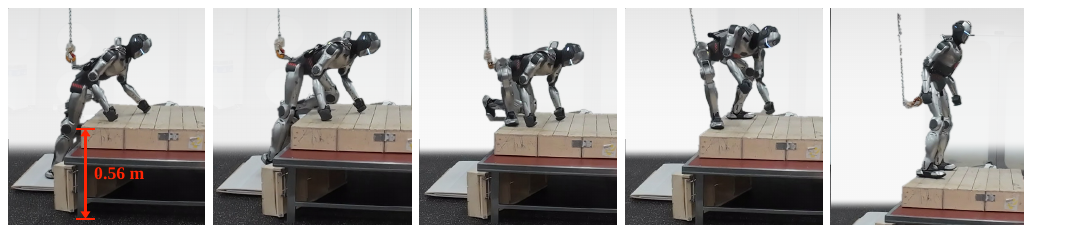}
        \label{fig:climb_high}
    }
    \caption{Real-world locomotion on unseen terrain without terrain observations: (a), (b) blind stair locomotion, and box climbing with a single policy onto (c) a platform 0.40~m high, which matches the reference motion, and (d) a platform 0.56~m high, which is 40\% higher than the reference.}
    \label{fig:variable_height_climbing}
\end{figure*}
\section{Real-World Experiments}
\label{sec:plane_real_world}
Our proposed framework enables reliable human-like execution, physical robustness, and task adaptation on hardware, even when neither refined reference motions nor detailed reward shaping is available.  
We trained all policies in this section with RSL-RL and IsaacLab, following the algorithm implementation and the domain-randomization and deployment settings in Section~\ref{sec:implementation_details}.

The tasks are divided into two groups serving different purposes.
For flat-terrain walking and blind stair locomotion, the task reward includes auxiliary style-shaping terms to ensure a fair comparison in a general locomotion setting.
With these terms, AMP learns both tasks but performs worse than PDM on each, and we quantify this difference for flat-terrain walking. The remaining tasks use a more challenging setting, combining kinematically retargeted reference motions with only core task rewards in Table~\ref{tab:rewards}. Under these conditions, AMP learns poorly in simulation and fails to achieve the task goals, as its style reward provides insufficient guidance, possibly due to vanishing gradients associated with discriminator saturation. PDM learns all tasks, and the resulting policies transfer successfully to hardware.

\begin{figure*}[t]
    \centering
    \subfloat[Getup]{
        \includegraphics[]{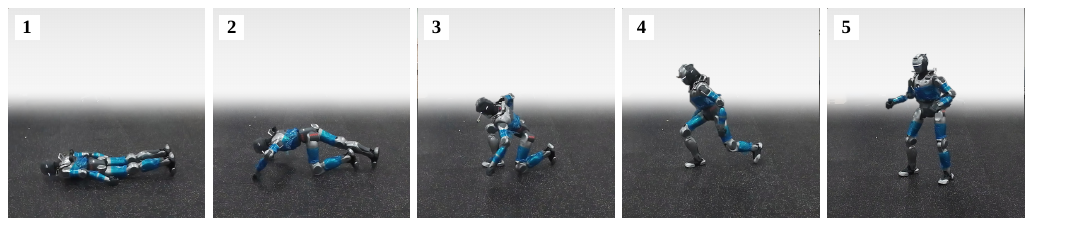}
    }\\
    \subfloat[Push Forward]{
        \includegraphics{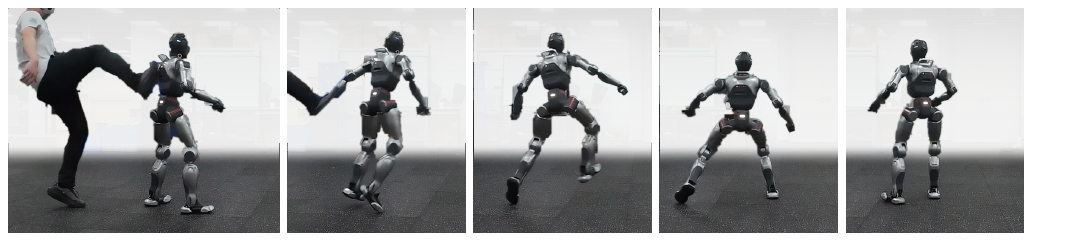}
    }\\
    \subfloat[Push Downward]{
        \includegraphics{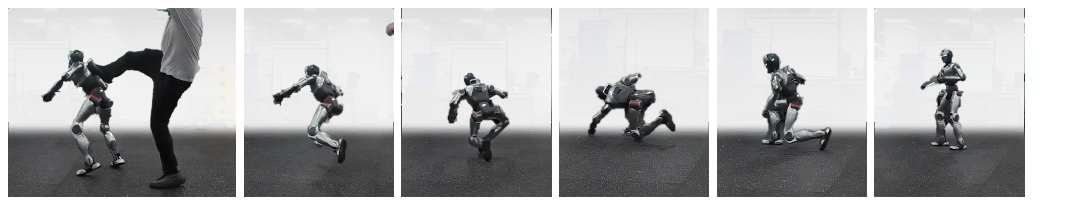}
    }
    \caption{Real-world get-up and recovery behavior. (a) The robot gets
    up from the ground. (b) After a forward push, it regains a stable
    stance by stepping. (c) After a downward push, it falls and then
    returns to standing.}
    \label{fig:perturbation_recovery}

\end{figure*}

\subsection{Style Preservation on Uneven Terrain}
\label{subsubsec:box_climbing}
Humans can navigate stairs or climb onto boxes carefully and
instinctively, even without vision. This capacity and the ability to
perceive an object's spatial properties through contact are of interest
in robotics.
Prior work uses predefined reference motions to reward robot exploration over terrain~\cite{wang2026spark} or distills a reference-guided policy into a policy conditioned on exteroceptive observations from additional sensors~\cite{wu2602perceptive}.
We treat the adaptation of behavior to terrain as the central problem and formulate it as distribution matching across terrains.
We evaluate this formulation on blind stair locomotion and box climbing.
For both tasks, the downstream policy is an MLP that handles terrain contacts through proprioceptive feedback alone, without access to terrain geometry, exteroceptive observations, or reference trajectories.

Blind stair locomotion builds on flat-terrain walking, with the expert additionally tracking stair-ascending and stair-descending trajectories with step heights of 15--20~cm.
Figure~\ref{fig:variable_height_climbing} shows that PDM generalizes the demonstrated skill to stair heights absent from the reference trajectories, enabling the deployed policy to traverse stairs without terrain observations.

Box climbing tests generalization to platform heights that differ from the reference.
The box-climbing task reward in Table~\ref{tab:rewards} encourages only
raising the base and placing the feet on the top surface. 
Figure~\ref{fig:variable_height_climbing} shows the same policy on a
platform 0.40~m high, matching the reference, and a platform 0.56~m high, which is
40\% higher.
On both platforms, the robot places its hands on the top surface, brings its legs onto it, and rises to a standing posture.
The policy preserves the reference's climbing structure while adapting
to the higher platform without observing its height. 
We also observed that the robot retried the step when its leg failed to reach the top surface. 
With this reward and the kinematically retargeted reference motion,
AMP did not achieve the climbing goal in simulation.

Basically, both tasks indicate that PDM remains applicable on challenging terrain using proprioceptive observations alone.
This is primarily because the distribution-matching reward provides a soft style constraint by rewarding transitions that resemble expert behavior, allowing the downstream policy to adapt to different terrains while preserving the desired motion style across the tested stair and platform heights.

\subsection{Style Preservation Under Large Deviations}
\label{subsubsec:perturbation_recovery}
Using a single motion clip, we trained the robot to get up from a face-down posture.
The recovery task reward in Table~\ref{tab:rewards} rewards only an upright trunk and stable standing without prescribing the recovery motion, whereas the discriminator constrains how the robot reaches these states. However, the AMP policy trained under the same conditions in simulation became
stuck in an unnatural configuration, bending its body and supporting it
with its head. This failure possibly resulted from retargeting artifacts
or exploitation of the task reward.

We find that the downstream policy trained on PDM produces natural behavior after large state deviations. 
Although the policy was trained only to get up, 
natural recovery behaviors also emerged when the robot was pushed. 
In the hardware tests, we vary the direction, timing, and intensity of pushes to displace the robot from its goal, while Fig.~\ref{fig:perturbation_recovery}(a) shows its default get-up behavior.
After a mild push (Fig.~\ref{fig:perturbation_recovery}(b)), the robot remains on its feet and regains a stable stance within a few steps.
After a strong push (Fig.~\ref{fig:perturbation_recovery}(c)), it falls to the ground and returns to standing through a kneeling posture.

The robot adapts its recovery to the push intensity, absorbing a small
push by stepping and switching to a get-up motion after a hard push
causes a fall. 
Both responses use coordinated, natural motions that the
task reward does not specify. 
These observations indicate that the distribution-matching reward constrains feasible behavior both near and far from the nominal trajectory.
Stronger motion matching may further reduce the need for manually
designed rewards and handcrafted components. 
This possibility warrants further study with matching-based approaches.

\begin{figure}[!t]
    \centering
    \includegraphics[trim=0bp 9bp 0bp 6.5bp, clip, width=\columnwidth]{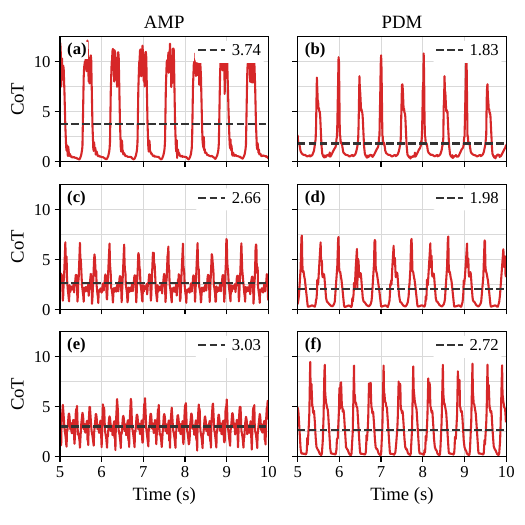}
    \caption{CoT of AMP (left column) and PDM (right column)
    at commanded forward velocities of 0.5, 1.0, and 1.5~m/s
    from top to bottom. Red solid curves represent instantaneous
    CoT, and black dashed lines indicate the mean over the
    displayed interval.}
    \label{fig:plane_cot}
\end{figure}

\subsection{Style Consistency Across Skills and Velocities}
We compared the AMP and PDM walking policies in simulation using their mechanical cost of transport (CoT) under different commands. 
Although CoT does not directly measure motion style, we use its time profile as indirect quantitative evidence, as it reflects gait regularity and periodicity.
Both policies were trained with the same task reward, including the auxiliary style-shaping terms.
The CoT at time $t$ is
\begin{equation}
    \mathrm{CoT}_t
    =
    \frac{\sum_{j}\big(\tau_{j,t}\,\dot q_{j,t}\big)^{+}}
         {m\,|v_{x,t}|},
    \label{eq:cot}
\end{equation}
where $\tau_{j,t}$ and $\dot q_{j,t}$ are the torque and velocity of joint $j$ at time $t$, $(\cdot)^{+}$ retains only positive mechanical power, $m$ is the robot mass, and $v_{x,t}$ is the forward velocity of the root.

Figure~\ref{fig:plane_cot} shows the CoT over a steady-state interval of 5~s during a simulated walking episode lasting 20~s.
At all three commanded velocities, the PDM trace consists of regular periodic peaks separated by low values.
The AMP trace, in contrast, shows broad high-CoT plateaus at 0.5~m/s and fluctuates between its main peaks at 1.0 and 1.5~m/s.
The mean CoT over the displayed interval is also lower for PDM at every velocity: 1.83, 1.98, and 2.72, compared with 3.74, 2.66, and 3.03 for AMP.
These results show that PDM maintains regular, periodic walking over the
tested velocity range and follows the walking commands with less
positive mechanical work than AMP.

In our experiments, PDM performed closely to the reference motion.
Figure~\ref{fig:plane} shows the deployed walking, two-leg jumping, and
crawling policies on flat terrain. Walking tests basic locomotion
stability, jumping involves dynamic transitions between takeoff and
landing, and crawling requires contact-rich motion in a nonstandard
body configuration.
We want to highlight that controllers for jumping and crawling can be trained simply by substituting the corresponding reference motions within the same PDM procedure of walking, without adding auxiliary style-shaping rewards.  
The distribution-matching reward makes the policy produce different motion styles and contact patterns, extending PDM beyond walking to other flat-terrain skills without manually designed gait-phase or leg-threshold rewards.
This improvement may arise from the ability of PDM to strengthen discriminator-based style guidance across different motion types by limiting discriminator reliance on motion artifacts.

\begin{figure}[!t]
    \centering
    \includegraphics[]{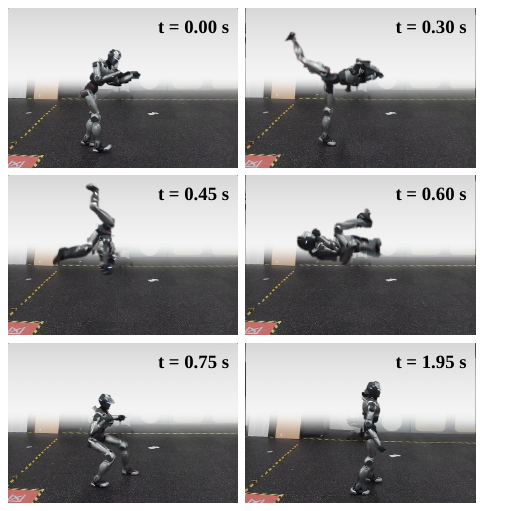}
    \caption{Snapshots of a real-world forward flip on the PM01 humanoid robot, from takeoff to stable standing.}
    \label{fig:flip}
\end{figure}
\subsection{Style Guidance for Agile Motions}
Figure~\ref{fig:flip} shows a forward flip executed on hardware.
A trigger command initiates the flip and is reset once the flip is completed.
The flip task reward in Table~\ref{tab:rewards} specifies trunk pitch
during rotation and a stable feet-first landing, without prescribing the
full-body motion. The downstream policy is sometimes prone to mode
collapse, for example, toward moving forward, and often omits key frames
of the reference trajectory.
This limitation is common to adversarial imitation methods.
With this minimal task reward and under extensive domain randomization
and action delay, PDM could still learn the flip, whereas AMP did not complete it
even with substantially longer training.
This result suggests that PDM's stronger distribution-matching reward
guides the robot toward the expert motion and may indirectly mitigate mode collapse.

\section{Limitations and Future Work}
\label{sec:limitations}
Stochastic rollouts from the tracking policy provide high-fidelity
motion samples for downstream training. Training both policies from
scratch requires separate PPO updates at each iteration, so an iteration
in our current implementation can take 1.5--2 times as long as an
iteration of single-policy training.
A pretrained tracking policy can instead generate the rollouts during
downstream training, moving the cost of its PPO updates to a separate
pretraining stage.
 
We evaluated the method on a limited set of real-world tasks using only
proprioceptive observations. More complex tasks in human environments,
such as carrying a box or climbing onto a high table, may require both
reference motions to guide behavior and exteroceptive feedback to adapt
to object geometry and location.
We will investigate how exteroceptive observations can support a broader
range of real-world tasks while preserving a human-like motion style.

We train a separate discriminator for each task using a small set of
task-specific reference motion clips. The downstream policy may still
exhibit mode collapse and reproduce only a narrow subset of the
behaviors in those clips.
Prior work has used latent representations to mitigate mode collapse~\cite{tirinzoni2025zero,li2026bfm,peng2022ase} and terrain conditioning to improve transitions across terrains~\cite{wang2025hil,chen2026lightlocoparkour}.
Building on these methods, we will test whether parallel-policy training
helps the robot perform multiple tasks within a shared command space
and combine learned skills.

\section{Conclusion}
\label{sec:conclusion}
This paper introduces PDM to address early discriminator saturation in adversarial imitation learning for legged robots.
Saturation can occur when the discriminator relies on a few features
that readily distinguish reference motions from policy rollouts.
To limit this reliance, we train the discriminator on stochastic rollouts
from both the tracking and downstream policies, using the same robot
model and physics simulator. In this setting, the discriminator's logit
approaches the equilibrium implied by its loss. The resulting distribution-matching
reward helps the downstream policy preserve the desired motion style
while achieving its task goals.

In simulation, physically feasible trajectories improved motion-tracking
accuracy and downstream policy performance but only slightly reduced
the discriminator logit gap. With stochastic rollouts, all input features
contributed to the discriminator's judgments of downstream samples.
Higher action noise reduced tracking accuracy, suggesting that it may
make the reward less informative for reproducing the expert motion.
In real-world experiments, the downstream policy performed tasks on
which the baselines struggled. It also maintained more consistent
motion patterns across the tested command range in both level-ground
and blind locomotion. The robot executed a flip in response to a trigger
command.

\section{Appendix}
\subsection{Derivation of the KL Term as a Return}
\label{app:kl_derivation}
This appendix derives \eqref{eq:kl_history} for the downstream policy
by means of the successor measure. For a set $X\subseteq\mathcal{S}$, let
the downstream successor measure be
\begin{equation}
    M_D(X\mid s)
    =
    \sum_{t=0}^{\infty}
    \gamma_D^t
    \Pr\bigl(s_{D,t}\in X \mid s_{D,0}=s,\pi_D\bigr),
\end{equation}
which accumulates the discounted probability that a rollout of $\pi_D$
from $s$ visits the set $X$. Because
$\sum_{t=0}^{\infty}\gamma_D^t=(1-\gamma_D)^{-1}$, the normalized
discounted visitation distribution is
\begin{equation}
    \rho_D(X)
    =
    (1-\gamma_D)\,
    \mathbb{E}_{s_0\sim\mu_D}
    \left[
        M_D(X\mid s_0)
    \right],
\end{equation}
or equivalently, for any function $g$,
\begin{equation}
    \mathbb{E}_{s\sim\rho_D}
    \left[
        g(s)
    \right]
    =
    (1-\gamma_D)
    \mathbb{E}_{\substack{
        \tau_D\sim\pi_D,\\
        s_0\sim\mu_D
    }}
    \left[
        \sum_{t=0}^{\infty}
        \gamma_D^t
        g(s_{D,t})
    \right].
    \label{eq:visitation_distribution}
\end{equation}
The expert successor measure $M_E$ and distribution $\rho_E$ are defined
analogously with $\pi_E$, $\gamma_E$, and $\mu_E$.

Substituting $\rho_D$ into the KL divergence gives
\begin{equation}
\begin{aligned}
    &\mathrm{KL}
    \left(
        \rho_D
        \,\Vert\,
        \rho_E
    \right)
    =
    \int_{\mathcal{S}}
    \log
    \frac{\rho_D(s)}{\rho_E(s)}
    \,\rho_D(ds)
    \\
    &\quad=
    (1-\gamma_D)\,
    \mathbb{E}_{s_0\sim\mu_D}
    \left[
        \int_{\mathcal{S}}
        \log
        \frac{\rho_D(s)}{\rho_E(s)}
        \,M_D(ds\mid s_0)
    \right]
    \\
    &\quad=
    -(1-\gamma_D)
    \mathbb{E}_{\substack{
        \tau_D\sim\pi_D,\\
        s_0\sim\mu_D
    }}
    \left[
        \sum_{t=0}^{\infty}
        \gamma_D^t
        \log
        \frac{\rho_E(s_{D,t})}{\rho_D(s_{D,t})}
    \right],
\end{aligned}
\label{eq:app_kl_downstream}
\end{equation}
which is \eqref{eq:kl_history}. The second equality uses the definition of $\rho_D$, and the third expands $M_D$ over the rollout. Minimizing the KL term thus maximizes the expected discounted return of the per-step reward
$\bar{r}_D(s)=\log\bigl(\rho_E(s)/\rho_D(s)\bigr)$ in
\eqref{eq:ideal_style_reward}.

Similarly, for the expert policy, $\mathrm{KL}(\rho_E\Vert\rho_D)$ gives
\begin{equation}
\begin{aligned}
    &\mathrm{KL}
    \left(
        \rho_E
        \,\Vert\,
        \rho_D
    \right)
    =
    \int_{\mathcal{S}}
    \log
    \frac{\rho_E(s)}{\rho_D(s)}
    \,\rho_E(ds)
    \\
    &\quad=
    (1-\gamma_E)\,
    \mathbb{E}_{s_0\sim\mu_E}
    \left[
        \int_{\mathcal{S}}
        \log
        \frac{\rho_E(s)}{\rho_D(s)}
        \,M_E(ds\mid s_0)
    \right]
    \\
    &\quad=
    -(1-\gamma_E)
    \mathbb{E}_{\substack{
        \tau_E\sim\pi_E,\\
        s_0\sim\mu_E
    }}
    \left[
        \sum_{t=0}^{\infty}
        \gamma_E^t
        \log
        \frac{\rho_D(s_{E,t})}{\rho_E(s_{E,t})}
    \right],
\end{aligned}
\label{eq:app_kl_expert}
\end{equation}
and the corresponding per-step reward is
$\bar{r}_E=\log\bigl(\rho_D/\rho_E\bigr)$.

Eqs.~\eqref{eq:app_kl_downstream} and \eqref{eq:app_kl_expert} differ in
three respects. First, each expectation is over rollouts of the policy in the first KL argument, so each reward is evaluated only on that policy's samples.
Second, the log ratio is inverted, so $\bar{r}_E(s)=-\bar{r}_D(s)$ at every state, and one discriminator can estimate both rewards with
$\bar{r}_D\approx z_{\varphi}$ and $\bar{r}_E\approx-z_{\varphi}$
(Section~\ref{sec:adversarial_matching}). Third, the normalizing factor $1-\gamma_E$ replaces $1-\gamma_D$ and is absorbed into $\beta_E$. Although $\bar{r}_E=-\bar{r}_D$ pointwise, the returns do not cancel in $\mathrm{KL}_\mathrm{sym}$ because their distributions differ.

\bibliographystyle{IEEEtran}
\bibliography{main}

\end{document}